\documentclass[journal]{IEEEtran}
\def\@IEEEclspkgerror{\ClassError{IEEEtran}}
\usepackage{enumitem}
\usepackage{cite}
\usepackage{graphicx}
\usepackage{algorithm}
\usepackage{algorithmic}
\usepackage{amsmath}
\usepackage{amssymb}
\usepackage{amsthm}

\usepackage{soul}
\usepackage[dvipsnames]{xcolor}
\usepackage{hyperref}
\usepackage[dvipsnames]{xcolor}

\makeatletter
\newcounter{parenttheorem}

\makeatother

\def\BibTeX{{\rm B\kern-.05em{\sc i\kern-.025em b}\kern-.08em
    T\kern-.1667em\lower.7ex\hbox{E}\kern-.125emX}}
\usepackage{lineno}
\begin{document}


\title{Adaptive Step Size Learning with Applications to Velocity Aided Inertial Navigation System}
\author{Barak Or, \IEEEmembership{Member, IEEE}, and Itzik Klein, \IEEEmembership{Senior Member, IEEE}
\thanks{Submitted: June 2022.}
\thanks{Barak Or and Itzik Klein are with the Hatter Department of Marine Technologies, Charney School of Marine Science, University of Haifa, Haifa, 3498838, Israel (e-mail: barakorr@gmail.com, kitzik@univ@haifa.ac.il).}}

\markboth{Adaptive Step Size Learning with Applications to Velocity Aided Inertial Navigation System / Or and Klein}%
{Or and Klein: Adaptive Step Size Learning with Applications to Velocity Aided Inertial Navigation System}
\maketitle

\begin{abstract}
Autonomous underwater vehicles (AUV) are commonly used in many underwater applications. Recently, the usage of multi-rotor unmanned autonomous vehicles (UAV) for marine applications is receiving more attention in the literature. Usually, both platforms employ an inertial navigation system (INS), and aiding sensors for an accurate navigation solution. In AUV navigation, Doppler velocity log (DVL) is mainly used to aid the INS, while for UAVs, it is common to use global navigation satellite systems (GNSS) receivers.  The fusion between the aiding sensor and the INS requires a definition of step size parameter in the estimation process. It is responsible for the solution frequency update and, eventually, its accuracy. The choice of the step size poses a tradeoff between computational load and navigation performance.  Generally, the aiding sensors update frequency is considered much slower compared to the INS operating frequency (hundreds Hertz). Such high rate is unnecessary for most platforms, specifically for low dynamics AUVs. 
In this work, a supervised machine learning based adaptive tuning scheme to select the proper INS step size is proposed. To that end, a velocity error bound is defined, allowing the INS/DVL or the INS/GNSS to act in a sub-optimal working conditions, and yet minimize the computational load. Results from simulations and field experiment show the benefits of using the proposed approach. In addition, the proposed framework can be applied to any other fusion scenarios between any type of sensors or platforms. 
\end{abstract}

\section{Introduction}
\label{sec:introduction}
Autonomous vehicles, such as Autonomous underwater vehicles (AUVs) or multi-rotor unmanned vehicles are commonly equipped with an inertial navigation system (INS) and other sensors \cite{farrell2008aided} to provide real-time information about their position, velocity, and orientation \cite{taheri2021slam,paull2013auv,huang2016multi,miller2010autonomous}. The INS has two types of inertial sensors, namely, the gyroscopes and accelerometers. The former measures the angular velocity vector, and the latter measures the specific force vector. Since the inertial sensors measurements contain noises, and error terms, the navigation solution drifts with time. In the underwater environment, usually, a Doppler velocity log (DVL) is used to reduce the solution drift with time \cite{karaboga2013adaptive,klein2018observability,tal2017inertial,elhaki2020robust}, while above the sea surface global navigation satellite systems (GNSS) measurements are used instead \cite{rogne2020usage,vasconcelos2011ins}. \\
The inertial sensors, DVL, and GNSS provide discrete information regarding a vehicle's continuous motion. Hence, tracking a vehicle involves a discrete realization of continuous motion. Such realization requires a step size selection, usually made by the designer according to the scenario and computational constraints  \cite{or2021kalman,shats1988discrete,kulikov2014accurate}. Moreover, to save power and extend the sensor/system life, the number of samples received from each source should be determined such as the information quality is maintained and the computational load is minimized \cite{dias2016adapting}. Most of the time, AUV navigate slow underwater for a long time. Thus, there is no need to obtain a navigation solution in a high frequency, except in situation of maneuvers where the drift might grow and there is a need for a momentary high computational load \cite{paull2013auv}. To cope with this trade-off an adaptive step size may be used.\\
In \cite{liu2008ekf}, an adaptive scheme for the step size, based only on the vehicle speed, was suggested for dealing with sensor scheduling in target tracking scenarios. An adaptive scheme was also suggested in \cite{ hamouda2010metadata}, where the step size is based only on the vehicle’s distance to the target. It improves the energy efficiency during target tracking scenarios. 
In \cite{or2021kalman}, a simple criterion was suggested to define the step size for sensor measurements to minimize computational load and still provide moderate navigation performance. This approach is based on the predictor and corrector of the linear discrete Kalman Filter (KF) \cite{kalman1961new,jazwinski2007stochastic}, where the main idea is to keep the discretized implementation of the continuous process with a lower numerical error. Later, an adaptive scheme to update the step size in real-time scenarios, with varying discrete noisy measurements was presented for constant velocity (CV) and constant acceleration (CA) models \cite{brown1992introduction}. \\
Focusing on inertial measurements step size, in a sensor fusion scenario like INS/DVL or INS/GNSS, the fusion is carried out using a nonlinear filter such as the Extended KF (EKF). There, inertial measurements are used in the system model while the aiding measurements are used in the filter measurement model to update the navigation state. The inertial sensors operate in a much faster frequency (tens or hundreds of Hertz) than the aiding sensors (several Hertz). As a consequence, approaches like in \cite{or2021kalman,dias2016adapting,liu2008ekf,hamouda2010metadata}, are not suitable for such setups, as they assume constant step sizes. \\
In order to avoid the need of constant step size, recent works explore the possibilities of using classical machine learning (ML) or deep learning (DL) based approaches \cite{goodfellow2016deep}. Such approaches recently lead to state of the art performance in various fields such as computer vision \cite{guo2016deep} and natural language processing \cite{cambria2014jumping}. Recently, ML and DL approaches are adopted in navigation applications. Among them are pedestrian dead reckoning where DL approaches are used to regress the user change in distance and heading \cite{yan2019ronin,yan2018ridi,klein2020stepnet,asraf2021,liu2020tlio}. In addition, DL based networks were used for attitude estimation using inertial sensors \cite{9119813,weber2020neural,vertzberger2021attitude} . ML approaches showed also better performance in stationary coarse alignment both in accuracy and time to converge \cite{zak2020mlca}. This trend of integrating ML approaches with classical INS applications raises the motivation to adapt such approaches also in the described problem of finding an adaptive step size. \\
In this area, a recent work by Dias et al. \cite{dias2016adapting}, discusses an adaptive step size of sensor networks, where online reinforcement learning technique is adopted to minimize the number of transmissions of the reported data. They showed a high reduction of energy while keeping the average information quality. However, that approach is limited to large step sizes and focused on a slow dynamics system, which is less relevant for navigation applications. \\
In this paper, a typical scenario of an adaptive step size determination for high rate inertial measurement unit (IMU) aided by a low-rate sensor such as DVL or GNSS is considered. There, the quality/amount-of-measurements trade-off to minimize velocity error as a function of the IMU step size is addressed in the EKF framework. In the proposed approach, ML models are used to predict the sub-optimal IMU step size, and handle the non-linearity of the INS model. Establishing a relationship between navigation features and their sub-optimal IMU step sizes can be applied in a real-time setting to solve the IMU step size conflict (accuracy vs. computational load). \\The main contributions of this paper are:
\begin{enumerate}
\item A numerical study of the effect of the inertial sensor step size on the velocity error in INS/DVL and INS/GNSS  typical fusion scenarios. 
\item Derivation of a learning-based scheme to determine an adaptive IMU step size as a function of the velocity error. 
\item Online integration of the proposed learning scheme with EKF implementation for the navigation filter.
\end{enumerate}

To validate the proposed approach two numerical examples of an AUV and a quadrotor are addressed, as well as quadcopter field experiments. Both simulations and experiments results show the benefits of implementing the proposed learning-based approach.\\
The rest of the paper is organized as follows: Section II deals with the problem formulation for INS/DVL and INS/GNSS models. Sections III presents the importance of step size selection, followed by novel learning-based step size tuning approach, where the feature engineering, database generation process, and adaptive tuning scheme with the INS are discussed. Section IV presents the simulations and field experiment results, and Section V gives the conclusions.
\section{Adaptive Navigation Filter}
\label{sec:Adaptive}
The nonlinear nature of the INS equations requires a nonlinear filter. The most common filter for fusing INS with external aiding sensors is the es-EKF \cite{farrell2008aided}. There, the errors are estimated and subtracted from the state vector. When considering velocity aided INS, the position vector is not observable \cite{klein2018observability}. Hence, it is not included in the error-state vector, defined as: 
\begin{equation}
\delta {\bf{x}} = {\left[ {\begin{array}{*{20}{c}}
{\delta {{\bf{v}}^n}}&{\delta {{\bf{\varepsilon }}^n}}&{{{\bf{b}}_a}}&{{{\bf{b}}_g}}
\end{array}} \right]^T} \in {{\mathbb{R}}^{12 \times 1}},
\end{equation}
where $\delta {{\bf{v}}^n} \in {{\mathbb{R}}^{3 \times 1}}$ is the velocity vector error states expressed in the navigation frame, ${{\bf{\delta \varepsilon }}^n} \in {{\mathbb{R}}^{3 \times 1}}$ is the misalignment vector expressed in the navigation frame, ${{\bf{b}}_a} \in {{\mathbb{R}}^{3 \times 1}}$ is the accelerometer bias residuals vector expressed in the body frame, and ${{\bf{b}}_g} \in {{\mathbb{R}}^{3 \times 1}}$ is the gyro bias residuals vector expressed in the body frame. The navigation frame is denoted by $n$ and the body frame is denoted by $b$. The navigation frame center is located at the body center of mass, where the $x$-axis points to the geodetic north, $z$-axis points down parallel to local vertical, and $y$-axis completes a right-handed orthogonal frame. The body frame center is located at the center of mass, $x$-axis is parallel to the longitudinal axis of symmetry of the vehicle pointing forwards, the $y$-axis points right, and the $z$-axis points down such that it forms a right-handed orthogonal frame. The linearized, error-state, continuous-time model is
\begin{equation} 
\delta {\bf{\dot x}} = {\bf{F}}\delta {\bf{x}} + {\bf{G}}\delta {\bf{w}},
\end{equation}
where $ {\bf{F}} \in {{\mathbb{R}}^{12 \times 12}}$ is the system matrix, ${\bf{G}} \in {{\mathbb{R}}^{12 \times 12}}$ is the shaping matrix, and $\delta {\bf{w}} = {\left[ {\begin{array}{*{20}{c}}
{{{\bf{w}}_a}}&{{{\bf{w}}_g}}&{{{\bf{w}}_{ab}}}&{{{\bf{w}}_{gb}}}
\end{array}} \right]^T} \in {{\mathbb{R}}^{12 \times 1}}$ is the system noise vector consisting of the accelerometer, gyro, and their biases random walk noises, respectively  \cite{bar2004estimation}. 
The system matrix, $\bf F$ and the shaping matrix $\bf G$ are provided in the appendix. We define ${{\bf{T}}_b^n}$ as the transformation matrix between body frame and navigation frame.
The corresponding discrete version of the navigation model (for small step sizes), as given in (2), is 
\begin{equation}
\delta {{\bf{x}}_{k + 1}} = {{\bf{\Phi }}_k}\delta {{\bf{x}}_k} + {{\bf{G}}_{{k}}}\delta {{\bf{w}}_k}.
\end{equation} 
The transition matrix, ${{\bf\Phi }_k}$, is defined by a first order approximation as 
\begin{equation}
{{\bf{\Phi }}_k} \buildrel \Delta \over = {\bf{I}} + {{\bf{F}}}\Delta {t},
\end{equation}
$k$ is a time index, $\delta {\bf{w}}_k$ is a zero mean white Gaussian noise, and ${\bf I}$ is an identity matrix. \\ The step size for the INS calculations is defined by 
\begin{equation}
\Delta t_k \buildrel \Delta \over = t_k-t_{k-1},
\end{equation}
where each step size is related to the IMU frequency, $\nu_{IMU}$, by
\begin{equation}
\Delta {t_k} = \frac{1}{{{\nu_{IMU}}}}.
\end{equation}
The discretized process noise is given by
\begin{equation}
{\bf{Q}}_k^d = {\bf{G}}{{\bf{Q}}^c}{{\bf{G}}^T}\Delta {t_k},
\end{equation}
where ${\bf Q}^c$ is the continuous process noise matrix.\\
The discrete es-EKF is used to fuse the INS with external measurements. The initial error state and error state covariance are defined as \cite{farrell2008aided,groves2015principles}
\begin{equation}
\begin{array}{*{20}{l}}
{\delta {{{\bf{\hat x}}}_0} = {{\bf 0}_{12 \times 1}}}\\
{{{\bf P}_0} = {{\bf{Q}}^d}},
\end{array}
\end{equation}
where $\delta {\bf \hat x}_0$ is the initial estimate error-state vector, and ${\bf P}_0$ is the initial covariance error. \\
The error-state vector is initialized every iteration, as following
\begin{equation}
\delta {\bf{\hat x}}_k^ -  = {\bf 0}.
\end{equation}
The error covariance propagation (prediction) is given by
\begin{equation}
{\bf{P}}_k^ -  = {{\bf{\Phi }}_{k - 1}}{\bf{P}}_{k - 1} {{\bf{\Phi }}_{k - 1}}^T + {\bf{Q}}_{k - 1}^d,
\end{equation}
where ${\bf P}_{k-1}$ is the estimate from previous state, $k-1$. The measurement arrives at time $j$, and then filter update is made. \\
The Kalman gain is given by 
\begin{equation}
{{\bf{K}}_j} = {\bf{P}}_j^ - {{\bf{H}}_j}^T{\left[ {{{\bf{H}}_j}{\bf{P}}_j^ - {{\bf{H}}_j}^T + {\bf{R}}_j^d} \right]^{ - 1}},
\end{equation}
where ${\bf R}^d$ is the discrete measurement noise covariance, assumed to be constant. \\
The error-state estimate update is given by
\begin{equation}
\delta {\bf{\hat x}}_j  = {{\bf{K}}_j}\delta {{\bf{z}}_j},
\end{equation}
where
\begin{equation}
\delta {{\bf{z}}_j} = {{\bf{\hat z}}_j} - {{\bf{z}}_j}\
\end{equation}
is the measurement residual vector, defined as the difference between the estimated (${\bf \hat z}_j$) and the actual (${\bf z}_j$) measurements.\\ 
Finally, the error covariance update (correction) is given by
\begin{equation}
{\bf{P}}_j  = \left[ {{\bf{I}} - {{\bf{K}}_j}{{\bf{H}}_{j}}} \right]{\bf{P}}_j^ -.
\end{equation}

\subsection{Velocity measurement models} 
Two types of velocity measurement models are considered: 1) DVL and 2) GNSS. Regardless of the measurement model, the velocity measurements are available in a constant frequency, with a different step size from the IMU. The step size of the aiding velocity sensor is given by
\begin{equation}
\Delta {{\bf{\tau }}_j} = {{\bf{\tau }}_j} - {{\bf{\tau }}_{j - 1}}, \,\,j= 1,2,...
\end{equation}
and related to the aiding sensor sampling frequency, $\nu_{Aiding}$, by
\begin{equation}
\Delta {{\bf{\tau }}_j} = \frac{1}{{{\nu_{Aiding}}}}.
\end{equation}
Without the loss of generality, it is assumed a constant step size for the aiding sensor measurements. Hence, the subscript $j$ is omitted for $\Delta \tau$. Commonly, the IMU has a higher frequency rate than the aiding sensor, thus, the following assumption is made:
\begin{equation}
\Delta {{\bf{\tau }}} \gg \Delta {t_k}{\rm{,}}\,\,\,\,\forall k.
\end{equation}
\subsubsection{DVL measurement model} 
After processing, DVL outputs the AUV velocity vector in the DVL frame, ${\bf{v}}_{DVL}^d$. Then, it is transformed to the body frame, ${\bf{v}}_{DVL}^b$, and eventually to the navigation frame, ${\bf{v}}_{DVL}^n$, where it is used in the navigation filter. Thus,
\begin{equation}
{\bf{v}}_{DVL}^n = {\bf{T}}_b^n{\bf{T}}_d^b{\bf{v}}_{DVL}^d,
\end{equation}
where ${\bf{T}}_d^b$ is the transformation matrix from the DVL frame to the body frame. For simplicity, it is assumed that ${\bf{T}}_d^b$ is accurately known,  and therefore, removed in further analysis. Linearizing (18) yields \cite{rothman2014analytical} 
\begin{equation}
\delta {\bf{v}}_{DVL}^b = {\bf{T}}_n^b\delta {\bf{v}}_{}^n - {\bf{T}}_n^b\left( {{{\bf{v}}^n} \times } \right)\delta {{\bf{\varepsilon }}^n}.
\end{equation}
The corresponding measurement residual is given by
\begin{equation}
\delta {\bf{z}}_{DVL,j}^b = {{\bf{H}}_{DVL,j}}\delta {{\bf{x}}_j} + {{\bf{\varsigma }}_{DVL,j}},
\end{equation}
where ${{\bf{\varsigma }}_{DVL,j}} \sim {\cal N}\left( {0,{{\bf{R}}_{DVL}^d}} \right) \in {{\mathbb{R}}^{3 \times 1}}$ is an additive discretized zero mean white Gaussian noise. It is assumed that ${{\bf{\varsigma }}_j}$ and $\delta {\bf w}_j$ are uncorrelated.
The corresponding time-variant DVL measurement matrix is given by
\begin{equation}
{\bf{H}}_{DVL,j} = \left[ {\begin{array}{*{20}{c}}
{{\bf{T}}_n^b}_j&{ - {{\bf{T}}_n^b}_j\left( {{{\bf{v}}^n} \times } \right)}_j&{{{\bf{0}}_{3 \times 6}}}
\end{array}} \right] \in {{\mathbb{R}}^{3 \times 12}}.
\end{equation}

\subsubsection{GNSS velocity measurement model}
The GNSS receiver outputs the velocity vector in the navigation frame. Hence, the corresponding time-invariant GNSS measurement matrix is given by
\begin{equation}
{\bf{H}}_{GNSS} = \left[ {\begin{array}{*{20}{c}}
{{{\bf{I}}_{3 \times 3}}}&{{{\bf{0}}_{3 \times 9}}}
\end{array}} \right] \in {{\mathbb{R}}^{3 \times 12}}.
\end{equation}
The corresponding measurement residual is given by
\begin{equation}
\delta {\bf{z}}_{_{GNSS},j}^b = {{\bf{H}}_{GNSS}}\delta {{\bf{x}}_j} + {{\bf{\varsigma }}_{_{GNSS},j}},
\end{equation}
where ${{\bf{\varsigma }}_{_{GNSS},j}} \sim {\cal N}\left( {0,{{\bf{R}}^d}_{GNSS}} \right) \in {{\mathbb{R}}^{3 \times 1}}$ is an additive discretized zero mean white Gaussian noise. It is assumed that ${{\bf{\varsigma }}_j}$ and $\delta {\bf w}_j$ are uncorrelated.

\section{Adaptive step size learning}
\subsection{Motivation: the importance of step size selection}
To demonstrate the influence of the step size on vehicle's velocity error, a  simplified simulated vehicle trajectory, shown in Figure 1, was used. The simulation parameters are summarized in Table I. As the example was conducted for a short period ($T=240[s]$), the IMU error model was simplified to include only zero mean white Gaussian noise:
\begin{equation}
{{{\bf{\bar f}}}_b} = {\bf{f}}_b^{imu} + {{\bf{w}}_a},
\end{equation}
and
\begin{equation}
{{{\bf{\bar \omega }}}_{ib}} = {\bf{\omega }}_{ib}^{imu} + {{\bf{w}}_g},
\end{equation}
where ${\bf{f}}_b^{imu}$ and ${\bf{\omega }}_{ib}^{imu}$ are true simulated outputs of the accelerometer and gyroscope, respectively. \\
In order to evaluate the navigation performance, Monte Carlo (MC) simulation with $100$ iterations was made. The averaged velocity root mean squared error (RMSE) for the entire scenario is $0.06 [m/s]$ which was obtained by setting $\Delta t=0.01[s]$.

\begin{table}[ht!]
\caption {INS/GNSS simulation parameters} \label{tab:title} 

\begin{center}
\begin{tabular}{ |c|c|c|c| } 
\hline
Description & Symbol & Value \\
\hline
GNSS velocity noise (var)        & $ R_{11},R_{22},R_{33}$      & $0.004^2 [m/s]^2$   \\ 
GNSS step size &$\Delta \tau$ & $1 [s]$   \\ 
Accelerometer noise (var) & $Q_{11},Q_{22},Q_{33}$   & $0.02^2[m/s^2]^2$      \\ 
Gyroscope noise (var) & $Q_{44},Q_{55},Q_{66}$   & $0.002^2[rad/s]^2$      \\ 
IMU step size & $\Delta t$ & $0.01[s]$      \\ 
Num. Monte Carlo iterations     & $N$           & $100$      \\ 
Simulation duration   & $T$   &$240 [s]$    \\
Initial velocity   & ${\bf{v}}^n_0$   &$[5, 0, 0]^T [m/s]$    \\
Initial position & ${\bf{p}}^n_0$   &$[32^0, 34^0,5[m]]^T $    \\

\hline
\end{tabular}
\end{center}
\end{table}

\begin{figure}[h!]
\centering
{\includegraphics[width=0.48\textwidth]{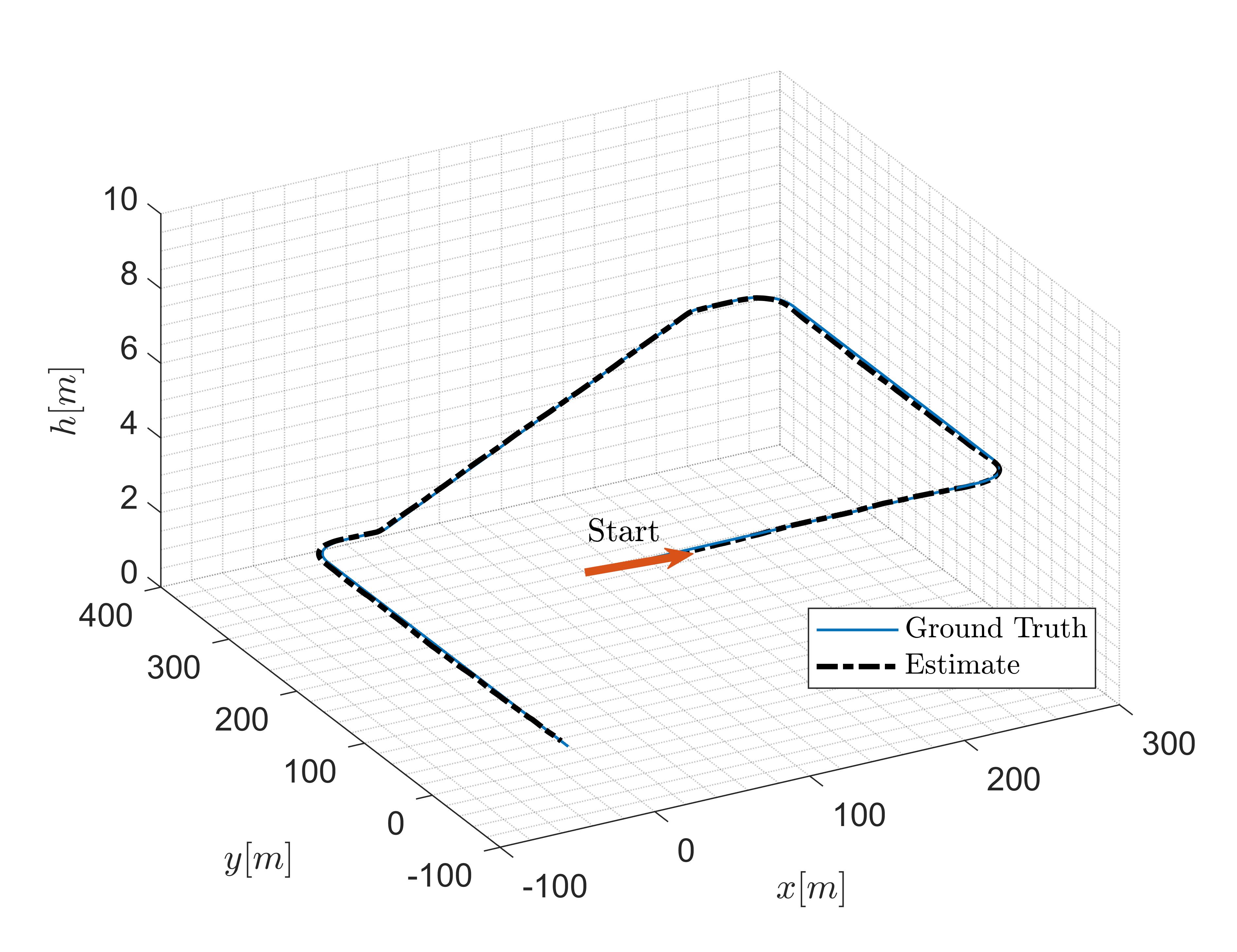}}
\caption{Estimated and ground truth simulated trajectories. Blue line shows ground truth and the black doted line shows the estimated trajectory. The plots are presented in a local Cartesian coordinate system.}
\label{[Fig1}
\end{figure}
As the velocity accuracy is affected by the predetermined step size of the IMU, a sub-optimal step size satisfies the following condition: 
\begin{equation}
\Delta {t^*} = \mathop {\arg \min }\limits_{\Delta t \in {\cal T}} \left[ {{\cal E}\left( {\Delta t} \right) - {\cal B}} \right]\,\,\,,{\mkern 1mu} {\mkern 1mu} {\mkern 1mu} {\cal B} > 0,
\end{equation}
where
\begin{equation}
{\cal E}{\rm{ }}\left( {\Delta t} \right) \buildrel \Delta \over = {\bf{E}}{\left\| {\delta {\bf{v}}_{True}^n\left( {\Delta t} \right)} \right\|_2}.
\end{equation}
The argument $\Delta t \in {\cal T} \subset \left[ {\Delta {t_{\min }},\Delta {t_{\max }}} \right]$ minimizes the difference between the $2^{nd}$ Euclidean norm of the mean averaged velocity (speed) error vector (${\cal E}$) and a design value, $\cal B$. Thus, the criterion allows velocity (speed) error up to  $\cal B$ for the sub-optimal step size. As $\cal B$ is a design parameter, one can choose it according to the platform and scenario at hand. In this work, we allow an averaged velocity error of 0.1[m/s]. \\
Obviously, when $\cal B$ goes to zero, $\Delta t \to 0$, and as a consequence the computational load increases. Hence, to find a trade off between accuracy and computational load, the condition, ${\cal B}>0$, must be satisfied. In due course, we performed 100 MC simulations each with 10 different IMU step sizes for three cases (different GNSS step size, $\Delta\tau$):
\begin{equation}
\Delta t \in \tilde {\cal T} = \left\{ {\begin{array}{*{20}{l}}
{0.002,0.004,0.008,0.01,0.016}\\
{0.02,0.032,0.04,0.05,0.1}.
\end{array}} \right\}
\end{equation}
The results are summarized in Figure 2, where each point represents the averaged velocity (speed) RMSE. For all scenarios, as the step size increases, the averaged velocity (speed) RMSE also increases. As seen in the figure, the change of the averaged velocity (speed) RMSE slowly converge to steady-state values for all step sizes per scenario. Hence, there is a $\Delta t$, which is too large to carry the navigation information, and it is associated with a high error. On the other hand, given a bound for the averaged velocity RMSE, $\cal B$, a moderate value of $\Delta t$ can be defined where the computational load will be minimized. For example, if $\Delta t=0.05[s]$ or $\Delta t=0.1[s]$ for case $[2]$ (Figure 2), are chosen, the same averaged velocity (speed) RMSE of $0.01[m/s]$ is achieved. For this example, $50\%$ of the computational load can be reduced without affecting the velocity error accuracy. Other cases consider the GNSS step size ($\Delta \tau$) as very small value (not necessarily available in the market, yet) - to demonstrate the impact of high-rate update. \\

\begin{figure}[h!]
\centering
{\includegraphics[width=0.48 \textwidth]{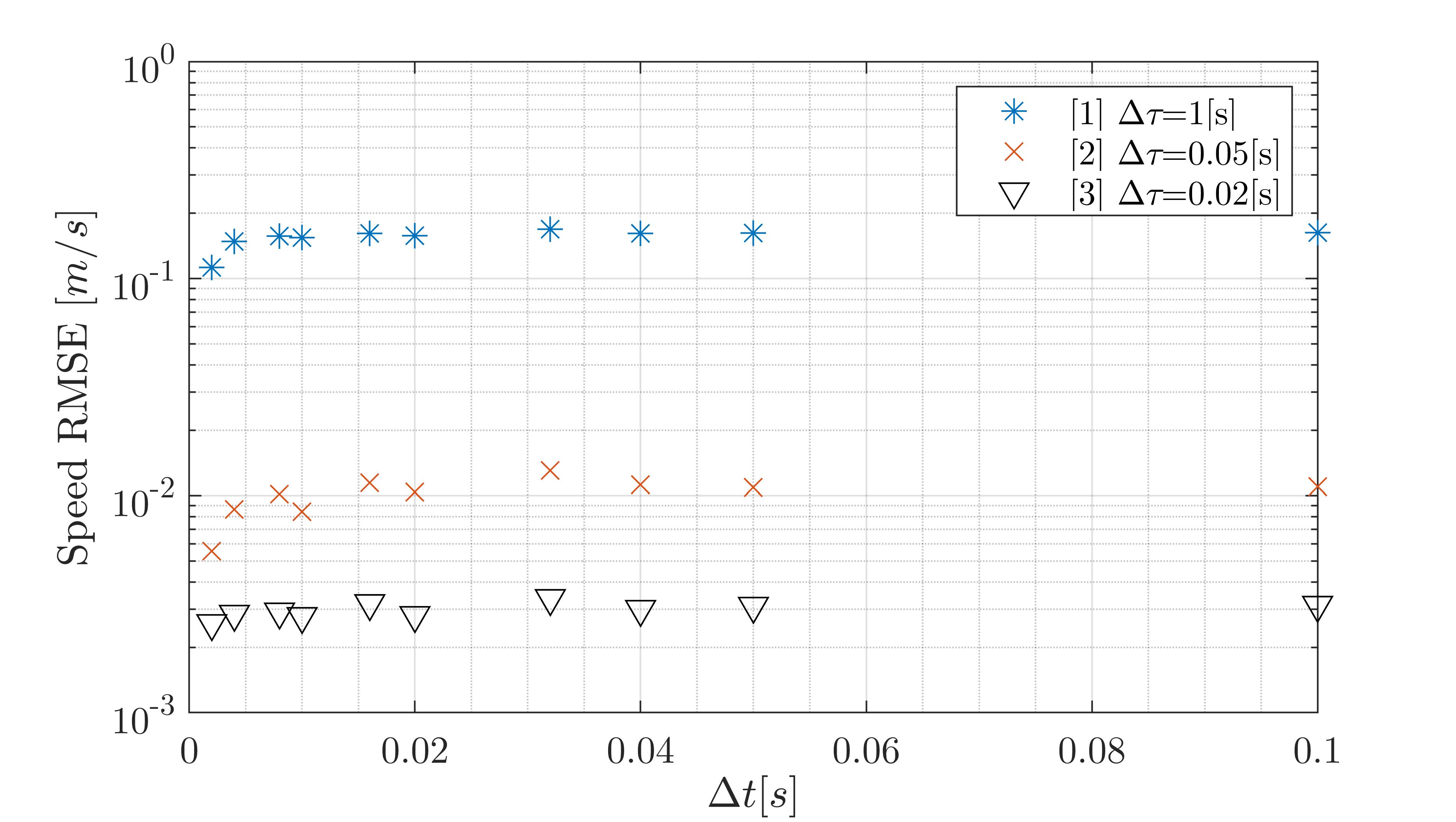}}
\caption{Velocity sensitivity to step size for five different scenarios with various GNSS step size ($\Delta \tau$). Complete trajectory and variances values are provided in Table 1.}
\label{[Fig2}
\end{figure}

\subsection{Supervised learning formulation}
The power of ML rises the ability to solve many difficult and non-conventional tasks. To determine the step size, the problem is formulated in a Supervised Learning (SL) approach; A feature set is defined where kinematic and statistics measures are considered. Formally, we search for a model to relate an instance space, ${\cal X}$, and a label space, ${\cal Y}$. We assume that there exists a target function, $\cal F$, such that ${\cal Y} = {\cal F}\left( {\cal X} \right)$. Generally, the SL task is to find ${\cal F}$, given a finite set of labeled instances:
\begin{equation}
\left\{ {{{\cal X}_k},\Delta t_k^*} \right\}_{k = 1}^N.
\end{equation}
The SL aims to find a function ${\tilde {\cal F}}$ that best estimates ${\cal F}$. A loss function, $l$, is defined to quantify the quality of ${\tilde {\cal F}}$ with respect to ${\cal F}$. The overall loss is given by
\begin{equation}
{\cal L}\left( {{\cal Y},\hat {\cal Y}\left( {\cal X} \right)} \right) \buildrel \Delta \over = \frac{1}{M}\sum\limits_{m = 1}^M {l{{\left( {y,\hat y} \right)}_m}},
\end{equation}
where $M$ is the number of examples, and $m$ in the example index. Minimizing $\cal L$ in a training/test procedure leads to the target function. The step size tuning problem is formulated as a classification problem, where only two classes are considered:
\begin{equation}
{\cal Y} = \left\{ {0.04,0.002} \right\} \in {{\mathbb{R}}^{1 \times 2}}.
\end{equation}
Ideally we would like to minimize the computational cost without influencing the accuracy. Yet,  due to the inherit tradeoff this is not possible. Therefore, we would like to minimize the computational cost with resulting minimum accuracy degradation. As a consequence, only the step-size is considered in the loss function, $l$, given by
\begin{equation}
l_m \buildrel \Delta \over = {\left( {\Delta t_m^* - \widehat {\Delta {t_m}}} \right)^2},
\end{equation}
where $\widehat {\Delta {t_m}}$ is the estimated step size value obtained by the learning model during the training process.
The main reasons for considering this problem as a classification task and not regression task are:
\begin{enumerate}
\item Filter robustness: Minimizing the amount of step size switching along the navigation process. 
\item ML model robustness: As there are only two classes in the label space. By doing so, the deterministic label space avoid invalid values and improve real-time performance. 
\end{enumerate}
Notice, that in $(30)$ two different step sizes are considered. Yet, if needed, the proposed framework can be applied for more different values pending on the scenario at hand. The major benefit of defining the problem as a bi-classification predictor, is that we minimize the number of “chattering” between many step size values (might lead to unstable filter). Also, using two values, one big $0.04[s]$ and the other small $0.002[s]$, presents clearly the computational effort reduction. Later, an example supporting the bi-classification choice is provided. 
\subsection{Feature engineering}
Sixteen high-level and low-level features, are considered: 
\begin{equation}
{\cal X} = \left\{ {{{\cal X}^{high\,}},{{\cal X}^{low}}} \right\} \in {{\mathbb{R}}^{1 \times 16}},
\end{equation}
where
\begin{enumerate}
\item High-level features:
A group of features that contains physical values of various filter and vehicle parameters in the scenario. Mean and square root are commonly used in many types of  classification/regression problems and thus used: 
\begin{equation}
{{\cal X}^{high\,}} = \left\{ \begin{array}{l}
\sqrt {{{\bf{Q}}_g}_{_{11}}} ,\sqrt {{{\bf{Q}}_a}_{_{11}}} ,\sqrt {{\bf{R}}_{^{11}}^d} ,\\
\Delta \tau ,{\bf{E}}{\left( {{{{\bf{\hat v}}}^n}} \right)^2},{\bf{E}}\hat \varphi^2 ,{\bf{E}}\hat \theta^2 ,{\bf{E}}\hat \psi^2 
\end{array} \right\} \in {{\mathbb{R}}^{1 \times 10}},
\end{equation}
where the subscript $_{11}$ stands for the first element of a matrix, and $\bf{E}$ is the expected value operator, calculated based on a moving average of the last 50 values. ${\bf{E}}{\left( {{{{\bf{\hat v}}}^n}} \right)^2}$ contains three features for north, east, and down velocity components.  $\phi,\theta,$ and $\psi$ are the body angles (briefly explained in the appendix). 
 \item Low-level features:
The low-level features are scalar values, created based on combination and modification of the high-level features, as summarized in Table II. Low-level features were chosen due to the dynamic body behavior and noise characteristics.
\end{enumerate} Generally, the designer can chose additional/different features for the classification task.

\begin{table}[ht!]
\caption {Low-level features} \label{tab:title} 
\begin{center}
\begin{tabular}{ |c|c|c|c| } 
\hline
Index & Feature & Description \\
\hline
${\cal X}_1^{low}$        & $\sqrt {{{\bf{Q}}_\omega}_{_{11}} + {{\bf{Q}}_{f_b}}_{_{11}} + {\bf{R}}_{^{11}}^d}$  &    Noise StD  2-norm      \\ 
${\cal X}_2^{low}$       &$\sqrt {\frac{1}{{{{\bf{Q}}_{\omega}}_{_{11}}}} + \frac{1}{{{{\bf{Q}}_{f_b}}_{_{11}}}} + \frac{1}{{{\bf{R}}_{^{11}}^d}}}$ &  Noise 'mean'     \\ 
${\cal X}_3^{low}$       & $\frac{{\sqrt {{{\bf{Q}}_{\omega}}_{_{11}}}  + \sqrt {{{\bf{Q}}_{f_b}}_{_{11}}} }}{{\sqrt {{\bf{R}}_{^{11}}^d} }}$ &    IMU + aiding sensor noise ratio  \\
${\cal X}_4^{low}$       & ${\bf{R}}_{^{11}}^d\Delta \tau$ &    Aiding sensor noise variance  \\
${\cal X}_5^{low}$       & ${\left\| {{{{\bf{\hat v}}}^n}} \right\|_2}$ &   Vehicle speed   \\
${\cal X}_6^{low}$       & $\frac{{\sqrt {{{\bf{Q}}_{\omega}}_{_{11}}}  + \sqrt {{{\bf{Q}}_{f_b}}_{_{11}}} }}{{\sqrt {{\bf{R}}_{^{11}}^d} }}{\left\| {{{{\bf{\hat v}}}^n}} \right\|_2}$ &   ${\cal X}_3^{low}* {\cal X}_5^{low} $\\
\hline
\end{tabular}
\end{center}
\end{table}

\subsection{Database generation process}
As a preliminary stage of training, a dataset should be generated. Then, it could be processed into the ML model. A velocity error, $\cal B$, was defined using different trajectories. There, the  vehicle traveled along them several times with various step sizes to find and store those that minimized vehicle velocity vector (speed) error. The various trajectories were created by modifying the radius of a circular motion, straight lines, and general curves. The process of generating such trajectories is divided into two parts: INS simulation with perfect IMU measurements to produce the GT trajectories and store them, and noisy velocity aided INS simulation in order to create noisy examples with their corresponding $\Delta t^*$ satisfying a desired bound of velocity error, $\cal B$. The IMU noise variance values as also velocity aiding sensor noise values are summarized in Table III.

\begin{table}[ht!]
\caption {Value ranges for Database generation} \label{tab:title} 

\begin{center}
\begin{tabular}{ |c|c|c| } 
\hline
Description & Value Range & Amount \\
\hline
Aiding velocity noise (var)  & ${{\bf{R}}_{ii}} \in {\left[ {0.0001,0.1} \right]^2}{\left[ {m/s} \right]^2}$ & $10$\\ 
Aiding velocity step size & $\Delta \tau  \in \left[ {0.1,2} \right]\left[ s \right]$ & $6$\\ Accelerometer noise (var) &${{\bf Q}_{ii}} \in {\left[ {0.0005,0.5} \right]^2}{\left[ {m/{s^2}} \right]^2}$& $10$ \\ 
Gyroscope noise (var) & ${{\bf Q}_{ii}} \in {\left[ {0.0001,0.1} \right]^2}{\left[ {rad/s} \right]^2}$ & $10$\\ 
\hline
\end{tabular}
\end{center}
\end{table}

The database generation process is described in Figure 3. IMU noise variances were firstly set, similarly to the simplified IMU error model provided in $(24)-(25)$ with different noise variances values. Then, IMU readings and aiding sensor measurements (constant step size) are processed into the velocity aided INS scheme and provide the vehicle navigation solution and their state errors as well. As the focus is to determine a sub-optimal $\Delta t$ to minimize velocity error, the error upper bound, $\cal{B}$, was chosen to be $0.1[m/s]$ (other values can be examined instead).
If the condition $(26)$ is satisfied, the example is stored. Else, the step size of the IMU, $\Delta t$, is reduced. Repeating this process for many scenarios yields a large dataset, enabling model training. Therefore, the examples in the dataset have the smallest step size, within the defined error bound, achieving the objective of trade-off balance between computation and accuracy
\begin{figure}[h!]
\centering
{\includegraphics[width=0.39\textwidth]{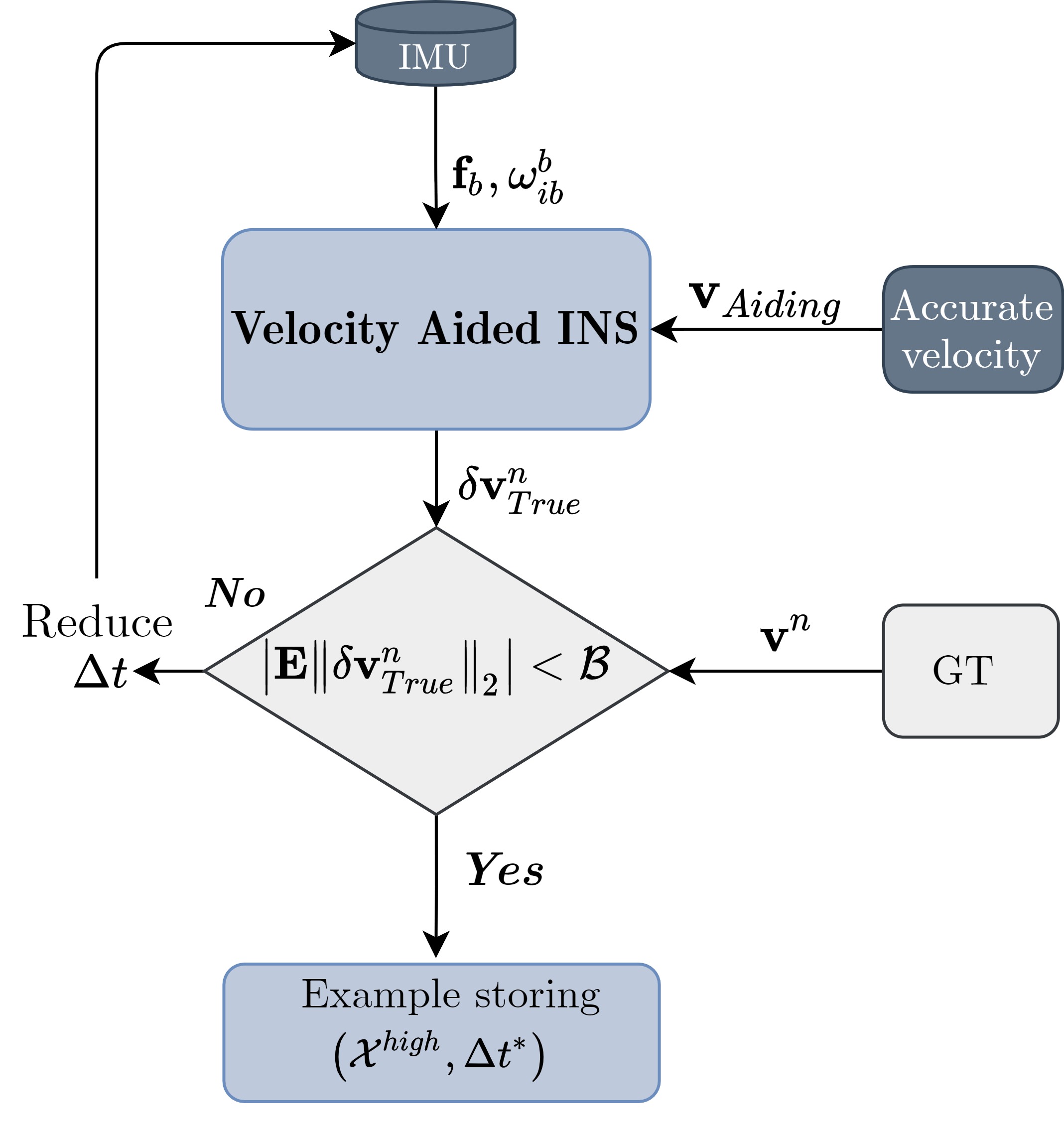}}
\caption{Database generation process. IMU and  accurate velocity measurements enter the velocity aided INS, then the velocity error is calculated. Given the GT velocity, the system decides if step-size should be reduced or not.}
\label{[Fig4}
\end{figure}

\subsection{Adaptive tuning scheme}
Applying the suggested tuning approach in online setting involves integration of the velocity aided INS with the classifier, as presented in Figure 4. Algorithm 1 gives the velocity aided INS with adaptive step size tuning algorithm. \\
\begin{algorithm}[H]
 \caption{Velocity aided INS with adaptive step size tuning}
 \begin{algorithmic}[1]
 \renewcommand{\algorithmicrequire}{\textbf{Input:}}
 \renewcommand{\algorithmicensure}{\textbf{Output:}}
 \REQUIRE ${\bf \omega}_{ib},{\bf f}_b,{\bf{v}}_{Aiding},\Delta t_0, \Delta \tau,T,tuningRate$
 \ENSURE  ${\bf{v}}^n,{\bf \varepsilon}^n$
 \\ \textit{Initialization} : ${\bf{v}}^n_0,{\bf{\varepsilon}}^n_0$
 \\ \textit{LOOP Process}
   \FOR {$t = 0$ to $T$}
   \STATE obtain ${\bf \omega}_{ib}$,${\bf f}_b$ 
  \STATE solve navigation equations (3)
  \IF {$(mod (t,\Delta \tau) = 0$)}
  \STATE obtain ${\bf{v}}_{Aiding}$ (20),(23)
 \STATE  update navigation state using the es-EKF (8)-(14)
  \ENDIF
  \STATE Calculate features and predict $\Delta t_{k+1}^*$. 
  \IF {$mod(t,tuningRate)=0$}
 \STATE calculate ${\cal X}$ (33)-(34)
\STATE$\Delta t_{k + 1}^* = {\tilde {\cal F}_{trained}}\left( {\cal X} \right)$
\ENDIF
  \ENDFOR
 \end{algorithmic}
 \end{algorithm}

\begin{figure}[h!]
\centering
{\includegraphics[width=0.4\textwidth]{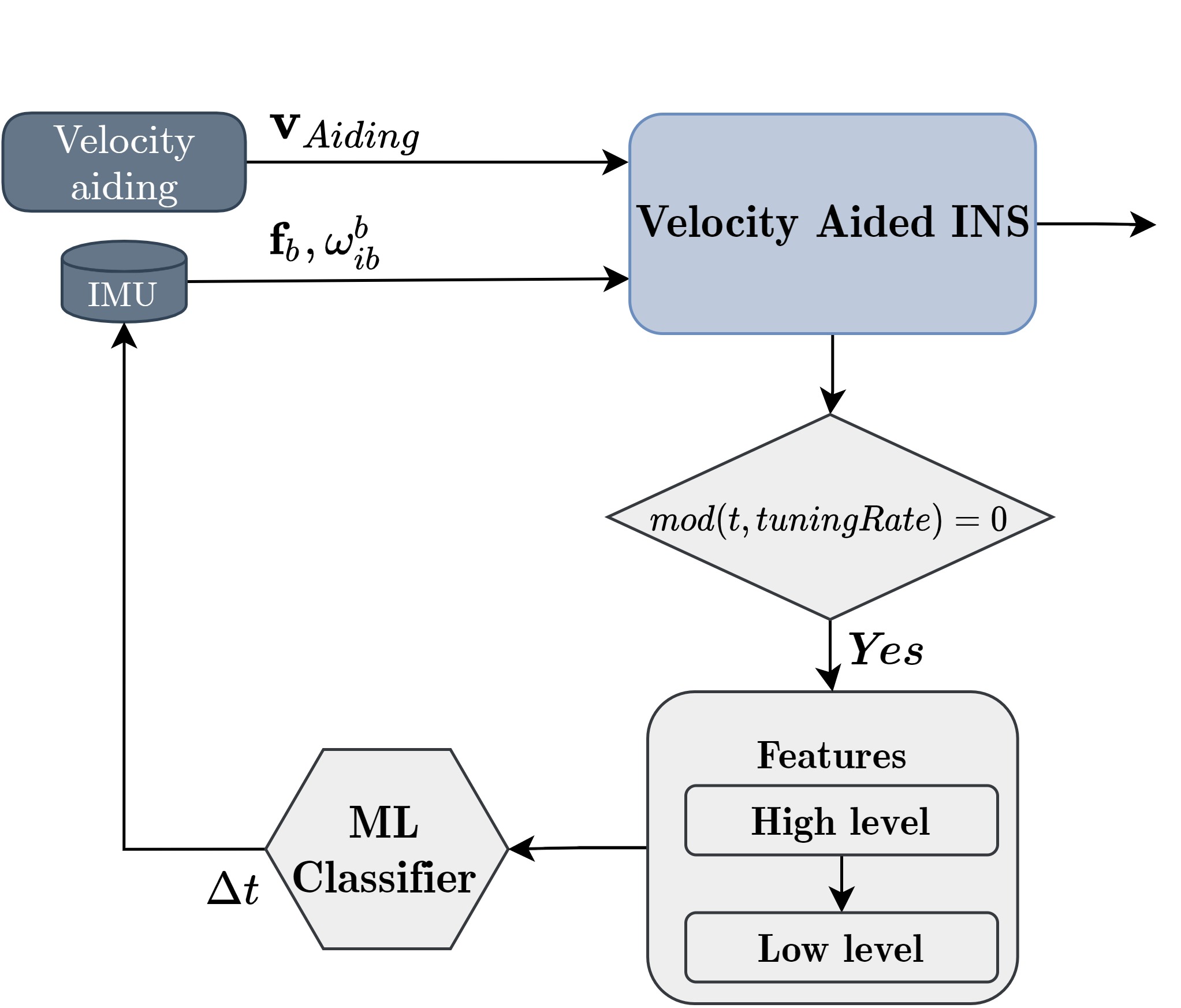}}
\caption{Step size adaptive tuning by applying the ML classifier. The features (high level and low level) are calculated based on temporal data. Then, they are processed into a classifier that outputs the sub-optimal step size.}
\label{[Fig6}
\end{figure}

\section{Results}
\subsection{Classification methods comparison}
In order to find the most suitable SL prediction model, various  classification models have been explored: fine tree (decision tree with many leaves), Naive Bayes, K-Nearest Neighborhood (KNN), Support Vector Machine (SVM), Logistic regression, and Ensemble (boosted trees) \cite{goodfellow2016deep,lecun2015deep}. The comparison process was made for 5 cases, with different IMU variance values, using a database (created as described in Section III.D) consisting of 18,000 examples. Those includes motion along straight lines with various velocities, and circles with different radiuses and velocities. All classifiers were bi-classifiers, with small and large step sizes for better model robustness (briefly explained in the appendix). The IMU, and aiding sensors noise covariances were tuned with different values to enrich the dataset. Note, that this dataset was created to handle with both INS/DVL and INS/GNSS fusion scenarios. \\
Two training paradigms were considered: train/test with 80/20 ratio, and cross validation with five folds. The vehicle's dynamics was set by tuning the IMU and selecting the initial kinematic conditions. \\ To evaluate the proposed models performance, the area under curve (AuC) measure was employed (see appendix for further explanations). For that, the positive value was defined as $P_{\Delta t}=0.002[s]$, and the negative value as $N_{\Delta t}=0.04[s]$. The second criterion used in order to evaluate the proposed models is the accuracy measure. Both criterions were calculated for each of the candidate models.\\
Classification results comparing ML approaches are provided in Figure 5. Each approach achieves AuC score and accuracy score for their classification performance, where once it was made by train/test paradigm and once by cross-validation paradigm. All models obtained more than 0.88 accuracy and AuC rates for both paradigms. The SVM obtained high accuracy rate ($0.95$) using the train/test paradigm, and high rate in the cross validation paradigm (also, a good performance according to the AuC rate). The Ensemble method slightly outperforms the SVM approach according to the AuC (both train/test and cross validation) as well as according to accuracy (cross validation). As the accuracy of both methods is similar, the SVM was chosen as, in general, it is known to be a robust classifier, as it maximizes the hyperplane margins \cite{cortes1995support}. Another justification to consider the SVM classifier is the trained model computational time. To that end, the excitation time was measured in the algorithm working environment (Intel i7-6700HQ CPU@2.6GHz 16GB RAM with MATLAB). The Ensemble model averaged iteration calculation time was 0.015[s] while the SVM was 0.001[s]. Hence, the SVM is 15 times faster than the Ensemble, which is a very important property in real time applications. These reasons lead us to choose the SVM as the optimal classifier for this task, as we deal with real-time scenarios and aim to keep the navigation filter robustness and efficient. The resulted ROC curve with a chosen classifier, $(FPR=0.03,TPR=0.87)$, received AuC of $0.98$, with accuracy of $0.95$ (obtained by a train/test paradigm). Hence, the SVM classifier was chosen for further analysis.  
\begin{figure}[h!]
\centering
{\includegraphics[width=0.48\textwidth]{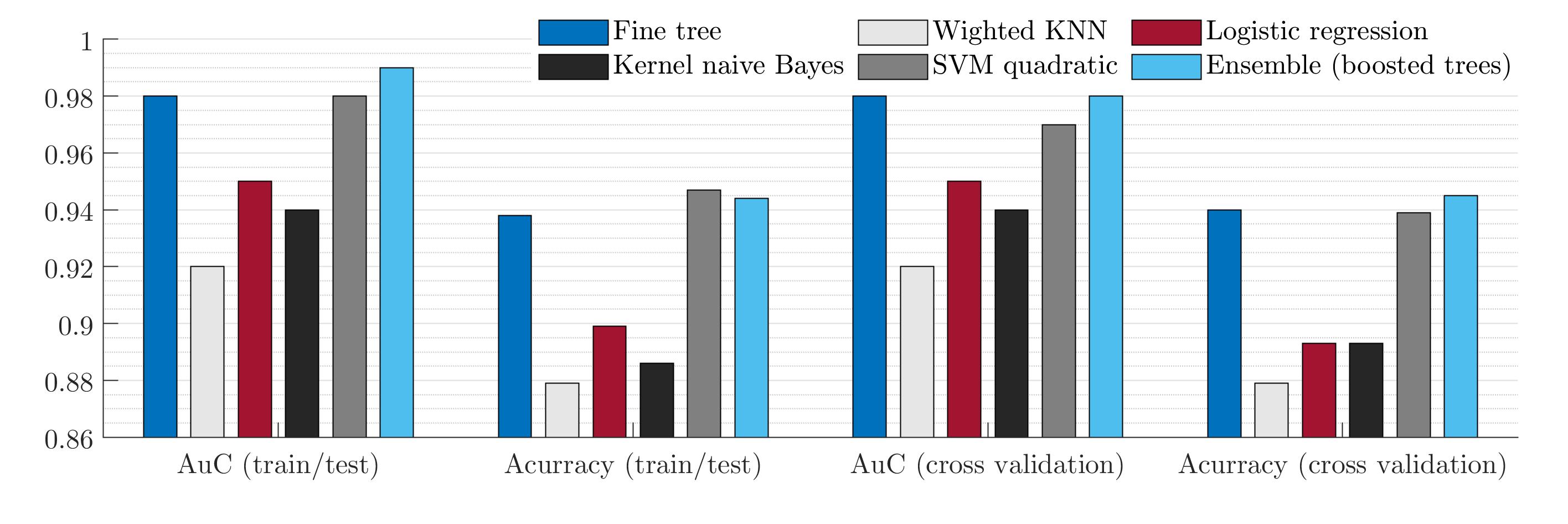}}
\caption{Learning model comparison with AuC and accuracy performance measures.}
\label{[Fig5}
\end{figure}

\subsection{MRMR based feature rank}
The learning-based models were trained using 16 features. In many real-time application the computational time of these features is critical and might take long time and eventually result in system latency or memory constraints. In order to avoid that, feature dimensionality reduction
 methods are applied. There, the features rank approach is used to select the most contributing features. One of the classical and common feature ranking approaches is the minimum redundancy maximum relevance (MRMR) method \cite{peng2005feature}. It was used to rank the feature set $(32)$. Figure 6 summarizes the results showing that ${\bf E}{\hat \psi}^2$, $\sqrt{{\bf Q}_{a_{11}}}$, and $\sqrt{{\bf R}_{11}^d}$are the three most important features. Although all features were used in the current research, the MRMR showed that the high level features contribute more to the classification process. Thus, if some computational constraint is required they could be removed from the analysis/model. 

\begin{figure}[h!]
\centering
{\includegraphics[width=0.48\textwidth]{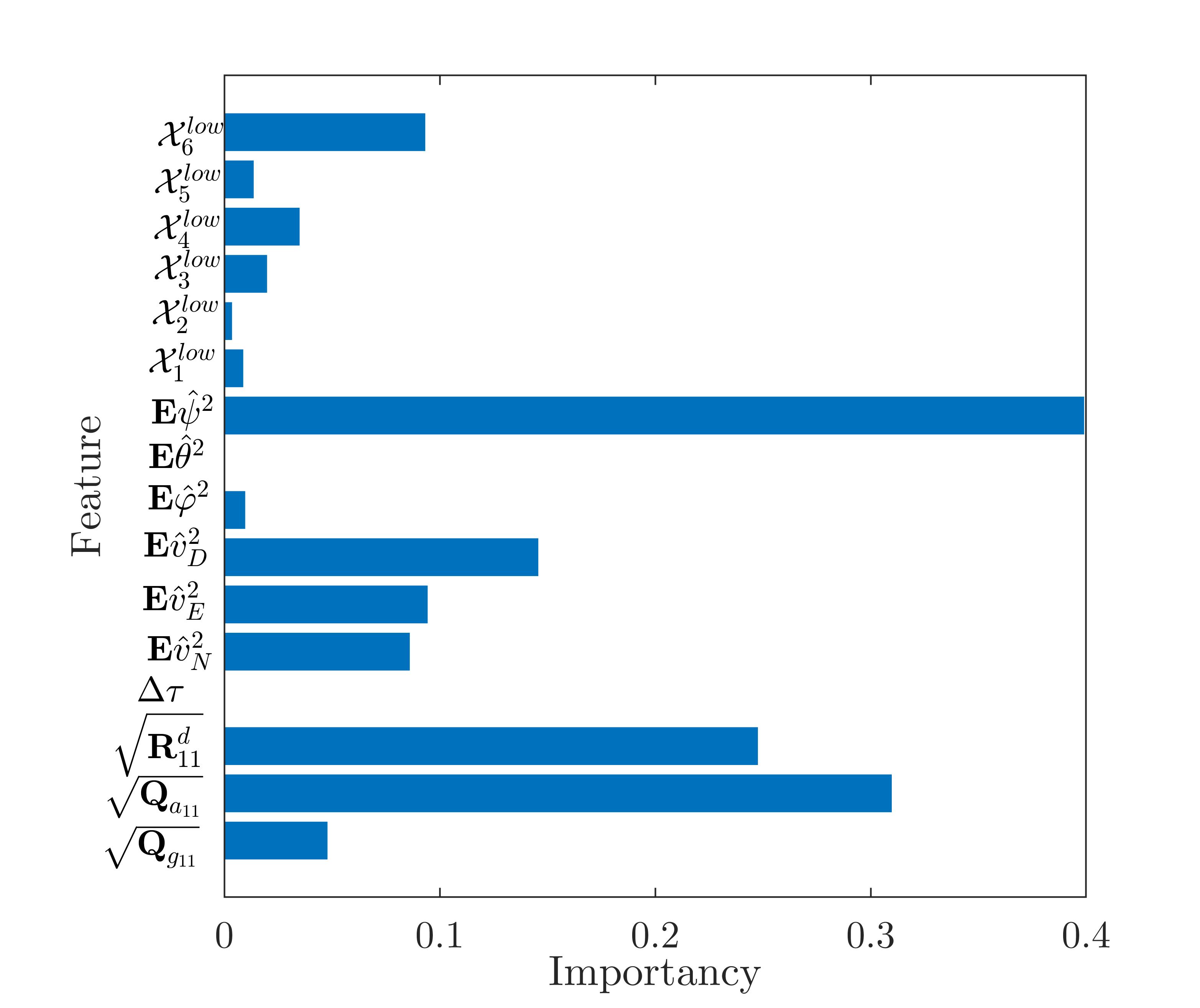}}
\caption{Feature importance ranking as obtained by the MRMR method.}
\label{[Fig7}
\end{figure}
\subsection{Simulations}
Two different simulation scenarios to validate the velocity aided INS with adaptive step size tuning were made. First, an INS/GNSS with scenario  parameters given in Table IV, and secondly, an INS/DVL with scenario parameters given in Table V. Both simulated trajectories are constructed by lines and curves, that the ML classifier was not trained on. However, it is expected the model will successfully capture the dynamics and statistics along the trajectories to predicting the sub-optimal $\Delta t^*$, as it was trained over $18,000$ examples (Section III.D) with various scenarios. For better comparison, we add one classical approach for step size tuning as a function of the velocity, given by \cite{liu2008ekf}:
\begin{equation}
\Delta t_{k + 1}^*\left( {{{\left\| {{{\bf{v}}^n}} \right\|}_2}} \right) = \left\{ {\begin{array}{*{20}{c}}
{\Delta {t_{\min }}}&{{{\left( {{{\left\| {{{\bf{v}}^n}} \right\|}_2}} \right)}_k} > {v^{Tresh}}}\\
{\Delta {t_{\max }}}&{{{\left( {{{\left\| {{{\bf{v}}^n}} \right\|}_2}} \right)}_k} \le {v^{Tresh}}}
\end{array}} \right.,
\end{equation}
where the velocity threshold, ${{v^{Tresh}}}$ is determined by the designer, upon the real-time scenario. 
In the adaptive setting, once the ML classifier predicts a different step size from the one it used in the last $20$ steps, the algorithm tunes the updated step size for the next iteration. \\
For the trajectory shown in Figure 1, graphs of the predicted $\Delta t$ as a function of time according to the ML classifier and the classical approach $(35)$ are plotted in Figure 7, for INS/GNSS simulation. There, the ML classifier predicted mostly $\Delta t^*=0.04[s]$, except for short time interval ($[50,58][s]$), where it predicted $\Delta t^*=0.002[s]$. 
The velocity error results are shown in Table VI, where, in addition to the adaptive tuning, two constant step sizes were used for comparison. Applying the small step size ($\Delta t=0.002[s]$) results in mean velocity error of $0.145[m/s]$, which is the lowest error associated with high computational load of $120,000$ iterations, and maximum velocity error of $0.41[m/s]$. From the other side, applying the larger step size ($\Delta t=0.04[s]$), used only $6,000$ iterations (only $5\%$ of the smaller step size) results in mean velocity error of $0.187 [m/s]$ (less than $0.05 [m/s]$ increase) and maximum velocity error of $0.655[m/s]$. By applying the adaptive step size tuning approach, a mean velocity error of $0.181[m/s]$ was obtained, with only $9,381$ iterations. This is less than $10\%$ of the conservative approach with IMU step size of $0.002[s]$, and also yields a lower velocity error than the large step size. The maximum velocity error with the adaptive step size is $0.37[m/s]$, lower than both  constant cases.\\ 
In Figure 8, the INS/DVL simulated trajectory is shown. The changes of $\Delta t^*$ during time according to the ML classifier and the classical approach $(35)$ are plotted in Figure 9. There, the ML classifier predicted $\Delta t^*=0.002[s]$ for the first $17[s]$ of the trajectory, and then predicted $\Delta t^*=0.04[s]$ until the end of the trajectory. 
The velocity error results are shown in Table VII, where, similarly to the INS/GNSS simulation, two constant step sizes were used to compare the adaptive tuning step size. Applying the smaller step size ($\Delta t=0.002[s]$) results in $0.015[m/s]$ mean velocity error and maximum velocity error of $0.019[m/s]$, which is the  error associated with high computational load of $20,000$ iterations. From the other side, applying the larger step size ($\Delta t=0.04[s]$), used only $1,000$ iterations, (only $5\%$ of the smaller step size) results in $0.0216 [m/s]$ mean velocity error (less than $0.07 [m/s]$ increase) and maximum velocity error of $0.046[m/s]$. By applying the adaptive step size tuning approach, a mean velocity error of $0.012 [m/s]$ was obtained with only $9,360$ iterations. This is less than a half from the conservative approach with IMU step size of $0.002[s]$, and also yields a lower mean velocity error. A maximum velocity error of $0.028[m/s]$ were obtained. For both INS/GNSS and INS/DVL simulations, the classical method for determining $\Delta t_k$ based on the vehicle speed, $(35)$, obtained insufficient computational load with higher mean velocity  error, where for the INS/GNSS a mean velocity error of $0.203[m/s]$ was obtained with over than $70,000$ iterations. and for the INS/DVL a mean velocity error of $0.280[m/s]$ was obtained with nearly $10,000$ iterations. The threshold was determined as the initial vehicles speed in both scenarios. To summarize,  while using our proposed adaptive approach, the average velocity error has increased by only $0.077[m/s]$ While using only about $1/7$ of the computational load.

\begin{table}[ht!]
\caption {INS/GNSS with adaptive step size simulation parameters.} \label{tab:title} 

\begin{center}
\begin{tabular}{ |c|c|c|c| } 
\hline
Description & Symbol & Value \\
\hline
GNSS velocity noise (var)        & $R_{11},R_{22},R_{33}$      &$0.02{[m/s]^2}$   \\ 
GNSS step size &$\Delta \tau$ & $1 [s]$   \\ 
Accelerometer noise (var) & $Q_{11},Q_{22},Q_{33}$   & $0.04^2[m/s^2]^2$      \\ Gyroscope noise (var)) & $Q_{44},Q_{55},Q_{66}$   & $0.003^2[rad/s]^2$ \\ 
Simulation duration   & $T$   &$240 [s]$    \\
Initial velocity   & ${\bf{v}}^n_0$   &$[5, 0, 0]^T [m/s]$    \\
Initial position   & ${\bf{p}}^n_0$   & $[32^0, 34^0,5[m]]^T $    \\
Velocity threshold   & ${v^{Tresh}}$   &$5[m/s]$    \\
\hline
\end{tabular}
\end{center}
\end{table}

\begin{table}[ht!]
\caption {INS/DVL with adaptive step size simulation parameters.} \label{tab:title} 

\begin{center}
\begin{tabular}{ |c|c|c|c| } 
\hline
Description & Symbol & Value \\
\hline
DVL noise (var)          & $R_{11},R_{22},R_{33}$      &$0.004{[m/s]^2}$   \\ 
DVL step size &$\Delta \tau$ & $1 [s]$   \\ 
Accelerometer noise (var) & $Q_{11},Q_{22},Q_{33}$   & $0.02^2[m/s^2]^2$      \\ Gyroscope noise (var) & $Q_{44},Q_{55},Q_{66}$   & $0.002^2[rad/s]^2$ \\ Simulation duration   & $T$   &$40 [s]$    \\
Initial velocity   & ${\bf{v}}^n_0$   &$[1, 0, 0]^T [m/s]$    \\
Initial position   & ${\bf{p}}^n_0$   & $[32^0, 34^0,-5[m]]^T $    \\
Velocity threshold   & ${v^{Tresh}}$   &$1[m/s]$    \\
\hline
\end{tabular}
\end{center}
\end{table}

\begin{table}[ht!]
\caption {INS/GNSS simulation results.} \label{tab:title} 

\begin{center}
\begin{tabular}{ |c|c|c|c|c| } 
\hline
$\Delta t[s]$ & Mean $\delta {\bf v}^n [m/s]$& Max $\delta {\bf v}^n [m/s]$ & Iterations \\
\hline
Adaptive (ours)        & $0.181$ &      $0.370$ &    $9,381$      \\ 
$0.002$       &$0.145$ & $0.410$ &  $120,000$     \\ 
$0.04$       & $0.187$ & $0.655$ & $6,000$  \\
$\Delta t\left( {{{\left\| {{{\bf{v}}^n}} \right\|}_2}} \right)$ & $0.203$ & $0.204$ & $70,410$ \\
\hline
\end{tabular}
\end{center}
\end{table}

\begin{figure}[h!]
\centering
{\includegraphics[width=0.47\textwidth]{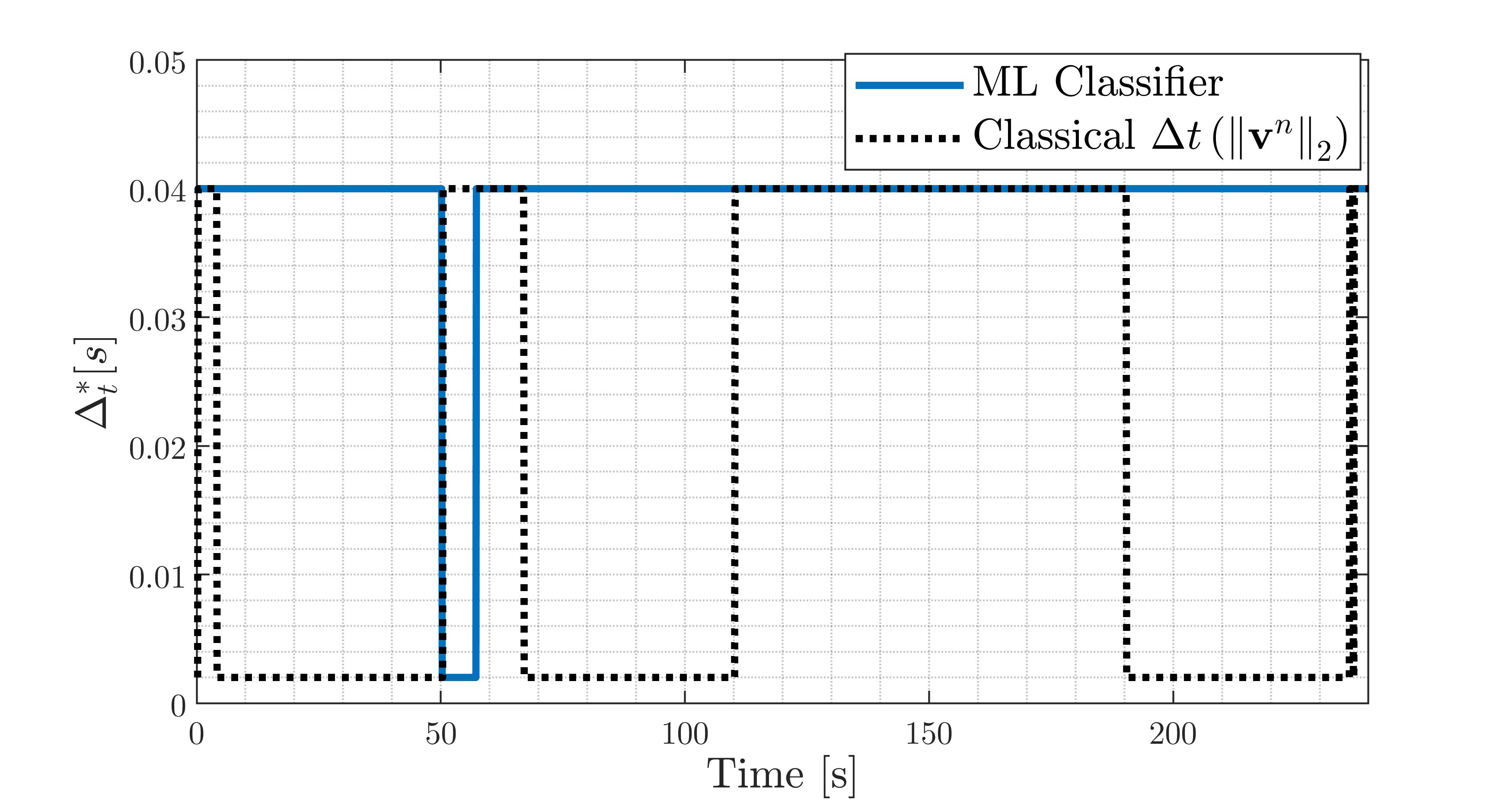}}
\caption{INS/GNSS simulation with sub-optimal step size, $\Delta t^*$, based on the ML classifier and classical $\Delta t\left( {{{\left\| {{{\bf{v}}^n}} \right\|}_2}} \right)$ as a function of time  for a duration of 4 minutes.}
\label{[Fig9}
\end{figure}

\begin{table}[ht!]
\caption {INS/DVL simulation results.} \label{tab:title} 

\begin{center}
\begin{tabular}{ |c|c|c|c| } 
\hline
$\Delta t[s]$ & Mean $\delta {\bf v}^n [m/s]$ & Max $\delta {\bf v}^n [m/s]$ & Iterations \\
\hline
Adaptive (ours)        & $0.012$  &   $0.028$  & $9,360$      \\ 
$0.002$   &$0.015$ & $0.019$& $20,000$     \\ 
$0.04$       & $0.0216$ &$0.046$ &    $1,000$  \\
$\Delta t\left( {{{\left\| {{{\bf{v}}^n}} \right\|}_2}} \right)$ & $0.0283$ & $0.076$ & $9,884$ \\
\hline
\end{tabular}
\end{center}
\end{table}

\begin{figure}[h!]
\centering
{\includegraphics[width=0.45\textwidth]{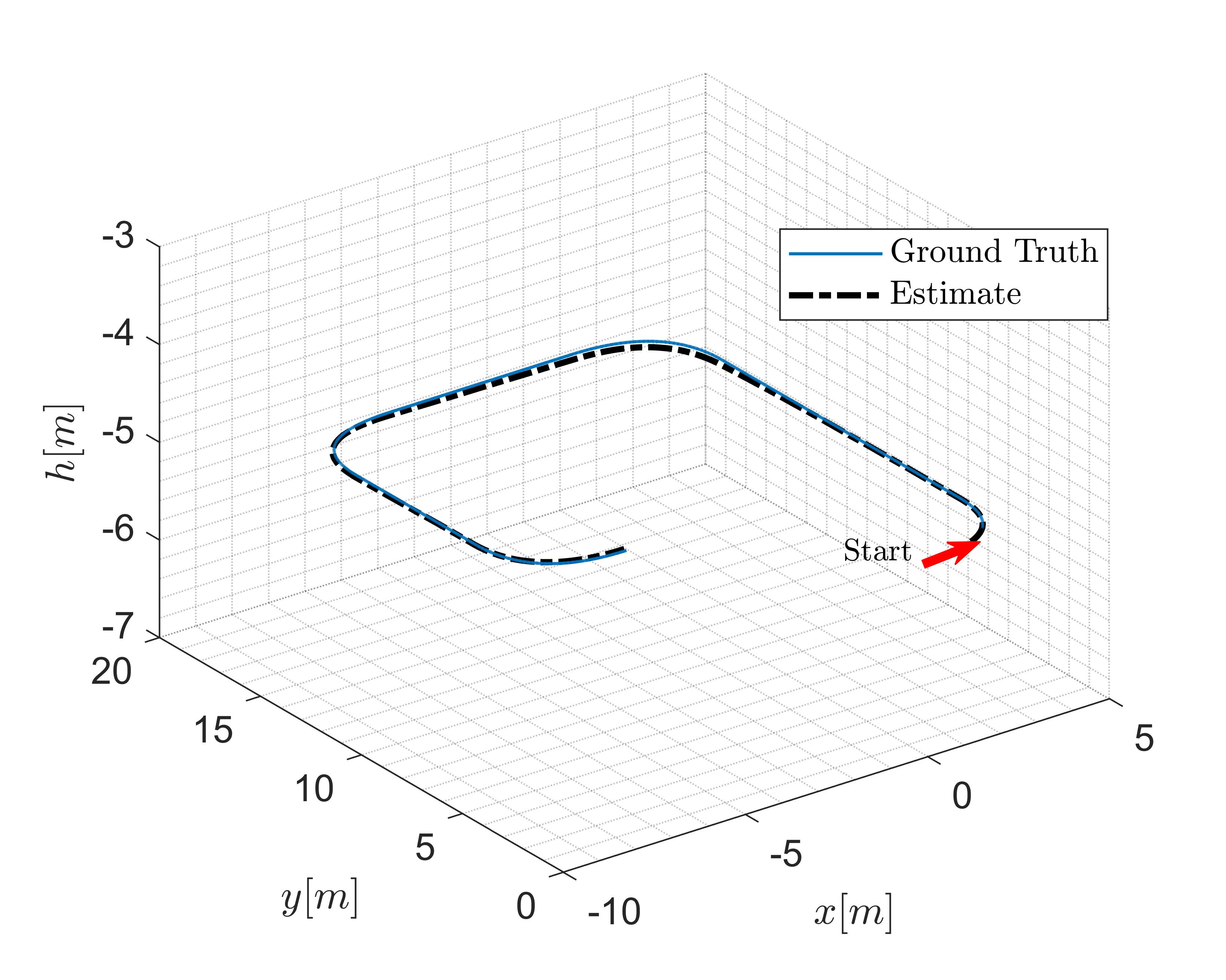}}
\caption{INS/DVL simulation trajectory for an AUV. The vehicle moves at the same sea level 
$(-5[m])$ and performs a rectangular motion. The blue line is for the GT trajectory and the dotted black line is for the estimated trajectory.}  
\label{[Fig11}
\end{figure}

\begin{figure}[h!]
\centering
{\includegraphics[width=0.45\textwidth]{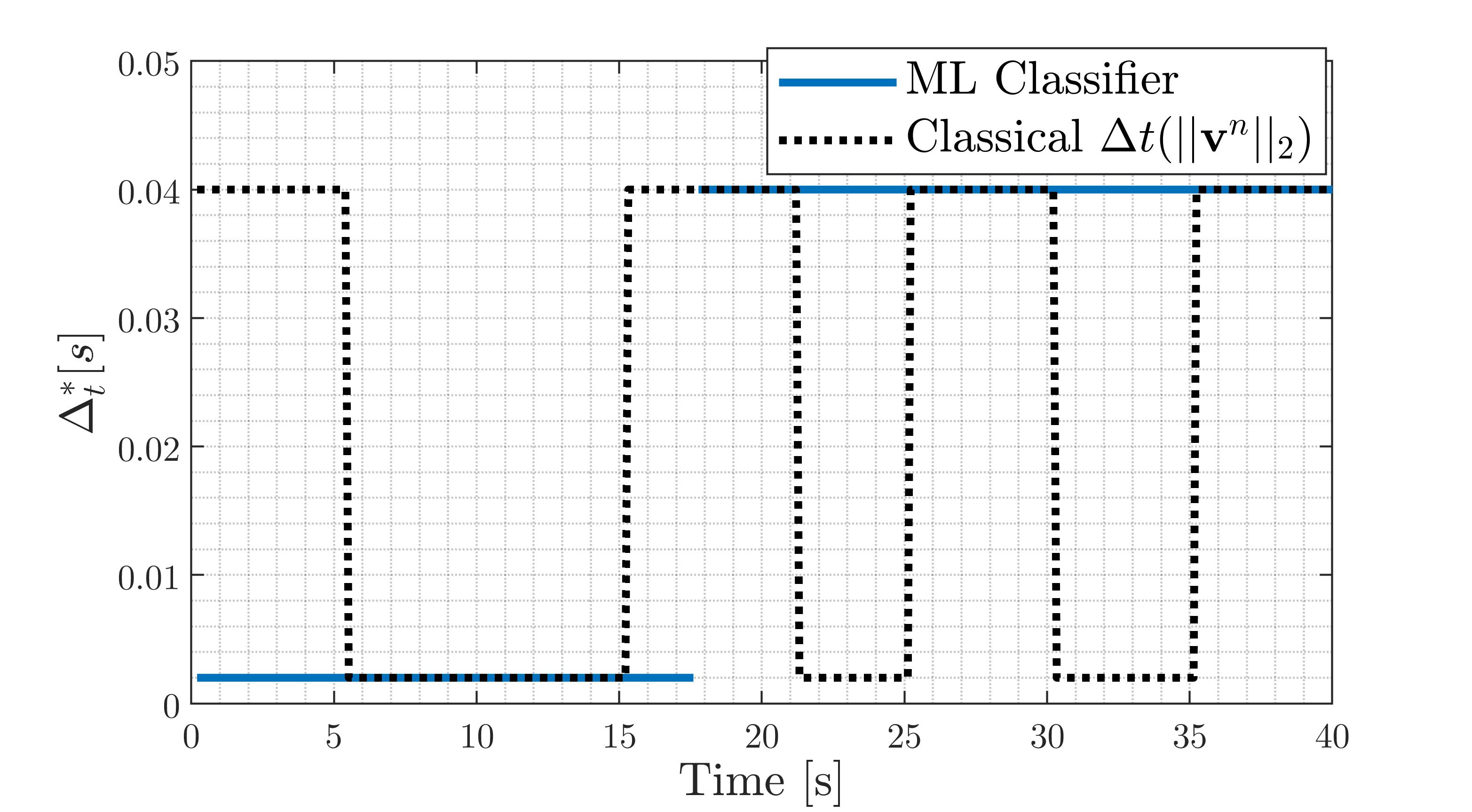}}
\caption{INS/DVL simulation with sub-optimal step size, $\Delta t^*$, based on the ML classifier and classical $\Delta t\left( {{{\left\| {{{\bf{v}}^n}} \right\|}_2}} \right)$ as a function of time  for a duration of 40 seconds.}
\label{[Fig10}
\end{figure}
\subsection{Field experiment}
A field experiment using a quadrotor was performed (Figure 10). The altitude of the quadrotor was kept constant and the horizontal trajectory is shown in Figure 11. The 12 error state model, described in Section II.A, was used to obtain the navigation solution. The filter was updated by velocity measurements from a GNSS receiver and the resulting navigation solution was compared with a GT measurements, obtained using an RTK device. To examine different (from the ones used in our simulations) and challenging scenarios (for the proposed ML algorithm), an “8-figure” shape trajectory was applied, to include accelerations/ declarations, as also turns for part of the time, and almost “straight” lines for the rest. Also, different parameters values were examined in the field experiment resulting in more cases that were examined and covered for better conclusion and generalization  of the ML approach, keeping the ML strategy the same. Experiment parameters are provided in Table VIII. In Algorithm 1, line 11, for the experiment, the optimal step size is $0.02[s]$. There is no need for additional training as the experimental dataset was addressed as a new test dataset, that is the ML was trained on the simulation training dataset. \\
The experiment error results using the classical approach,  $\Delta t_k=0.002[s]$, $\Delta t_k=0.02[s]$, and $\Delta t^*$ from the suggested adaptive tuning approach are summarized in Table IX. It appears that $1800$ iterations results in $0.128 [m/s]$ mean velocity error, and increasing the number of iterations to $18,000$ yields a meaningful reduction where only $0.01[m/s]$ mean velocity error is obtained. A maximum velocity error of $6.25[m/s]$ was obtained for a fractional initialization moment, both for the adaptive step size and the constant step size of $\Delta t_k=0.002[s]$. Our suggested adaptive tuning algorithm founds a sub-optimal solution, where only 9,900 iterations yields a $0.02[m/s]$ mean velocity error (as defined by setting ${\cal B}$). The designer controls the amount of iterations according to velocity RMSE criterion. In this experiment, the number of iteration using $\Delta t_k=0.02[s]$ increases adaptive by a factor of almost 6 (from $1800$ to $9900)$ to meet designer's criterion. From the other side, applying the adaptive tuning scheme results in $45\%$ reduction of number of iterations obtained while setting $\Delta t_k=0.002[s]$ with only $0.01[m/s]$ increase of mean velocity error. The maximum velocity error is $5.82[m/s]$ for a fractional initialization moment. In this experiment we obtained a nearly linear relationship between the velocity mean error and the step size. This result confirms our numerical simulation from III.A. The classical method for determining $\Delta t_k$ based on the vehicle speed, $(35)$, obtained insufficient computational load with higher mean velocity  error: a mean velocity error of $0.037[m/s]$ was obtained with $4,500$ iterations. The changes of $\Delta t^*$ during time according to the ML classifier and the classical approach (35) are plotted in Figure 12. 

\begin{figure}[h]
\centering
{\includegraphics[width=0.45\textwidth]{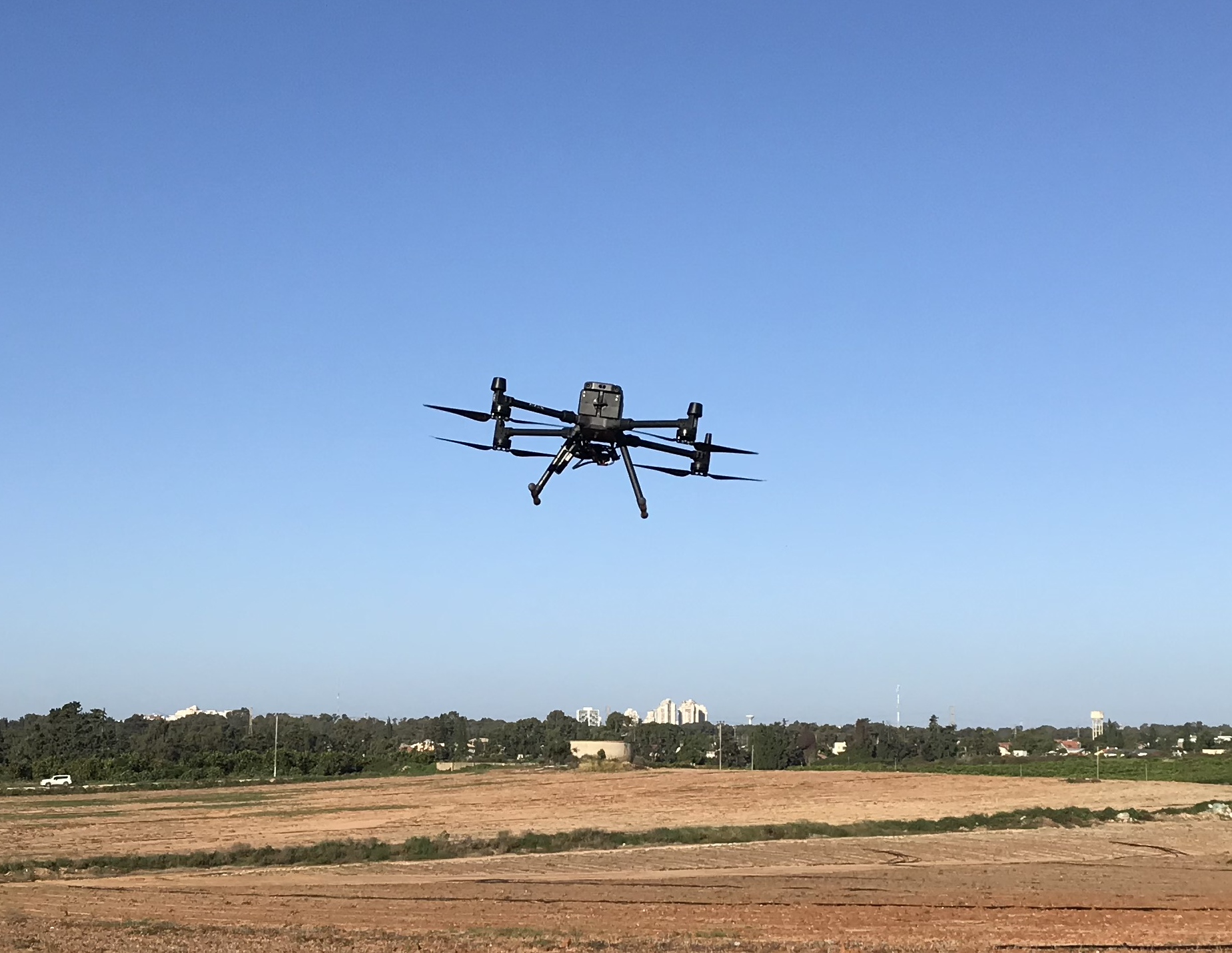}}
\caption{DJI matrice 300 in a field experiment.}
\label{[Fig11}
\end{figure}

\begin{figure}[h]
\centering
{\includegraphics[width=0.45\textwidth]{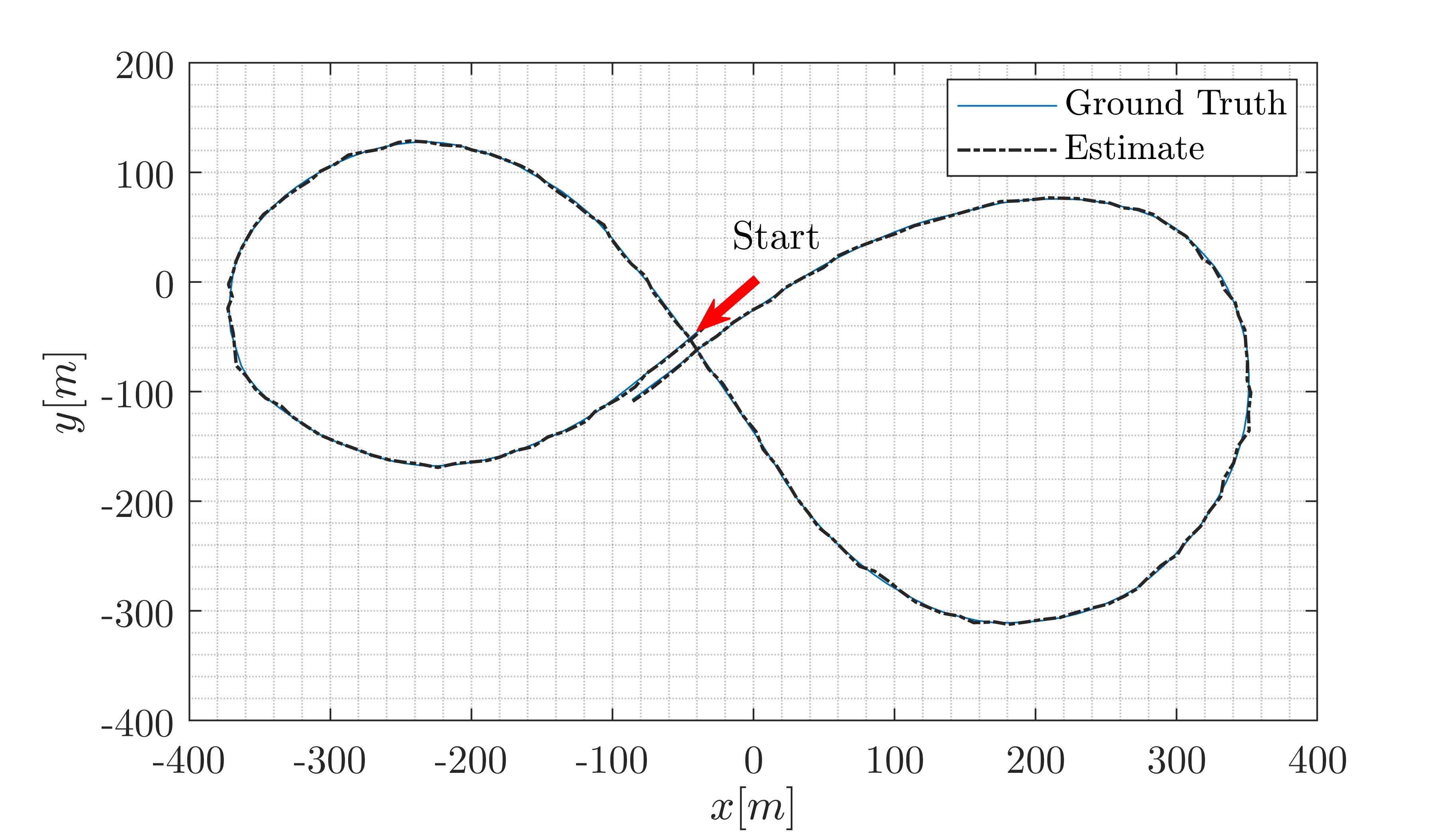}}
\caption{Field experiment Quadrotor GT trajectory and its estimated one.}
\label{[Fig12}
\end{figure}

\begin{table}[ht!]
\caption {Experiment parameters} \label{tab:title} 

\begin{center}
\begin{tabular}{ |c|c|c|c| } 
\hline
Description & Symbol & Value \\
\hline
GNSS vel. noise (var) - horizon.        & $R_{11},R_{22}$      &$0.003[m/s]^2$   \\ 
GNSS vel. noise (var) - vertical        & $R_{33}$      &$0.005[m/s]^2$   \\ 
GNSS step size &$\Delta \tau$ & $0.1 [s]$   \\ 
Accelerometer noise (MPU-9250) & $Q_{11}^*,Q_{22}^*,Q_{33}^*$   & $	
300[\mu g/\sqrt{Hz}]$      \\ 
Gyroscope rate noise (MPU-9250) & $Q_{44}^*,Q_{55}^*,Q_{66}^*$   & $0.01[dps/\sqrt{Hz}]$      \\ 
Experiment duration   & $T$   &$35 [s]$    \\
Initial velocity   & ${\bf{v}}^n_0$   &$[0, 0, 0]^T [m/s]$    \\
Initial position   & ${\bf{p}}^n_0$   &$[32.1^0, 34.8^0, 0]^T$    \\
Accelerometer bias noise &$Q_{7}^*,Q_{8}^*,Q_{9}^*$ & $[1,1,1] [m/s^2]^2$ \\
Gyroscope bias noise &$Q_{10}^*,Q_{11}^*,Q_{12}^*$ &$[1,1,1] [rad/s]^2$ \\
velocity threshold   & ${v^{Tresh}}$   &$7[m/s]$    \\
\hline
\end{tabular}
\end{center}
\end{table}

\begin{table}[ht!]
\caption {Errors as a function of step size \\ and the resulting number of Iterations} \label{tab:title} 
\begin{center}
\begin{tabular}{ |c|c|c|c| } 
\hline
$\Delta t_k$ [$s$] & Mean $\delta {\bf v}^n[m/s]$ & Max $\delta {\bf v}^n [m/s]$ & Iterations \\
\hline
Adaptive (ours)  & $0.0217$ &$6.25$& 9900  \\ 
0.002   & $0.0111$ &$6.25$& 18000 \\ 
0.02    & $0.1280$ &$5.82$& 1800  \\ 
$\Delta t\left( {{{\left\| {{{\bf{v}}^n}} \right\|}_2}} \right)$ & $0.037$ & $5.82$ & $4,500$ \\
\hline
\end{tabular}
\end{center}
\end{table}

\begin{figure}[h!]
\centering
{\includegraphics[width=0.45\textwidth]{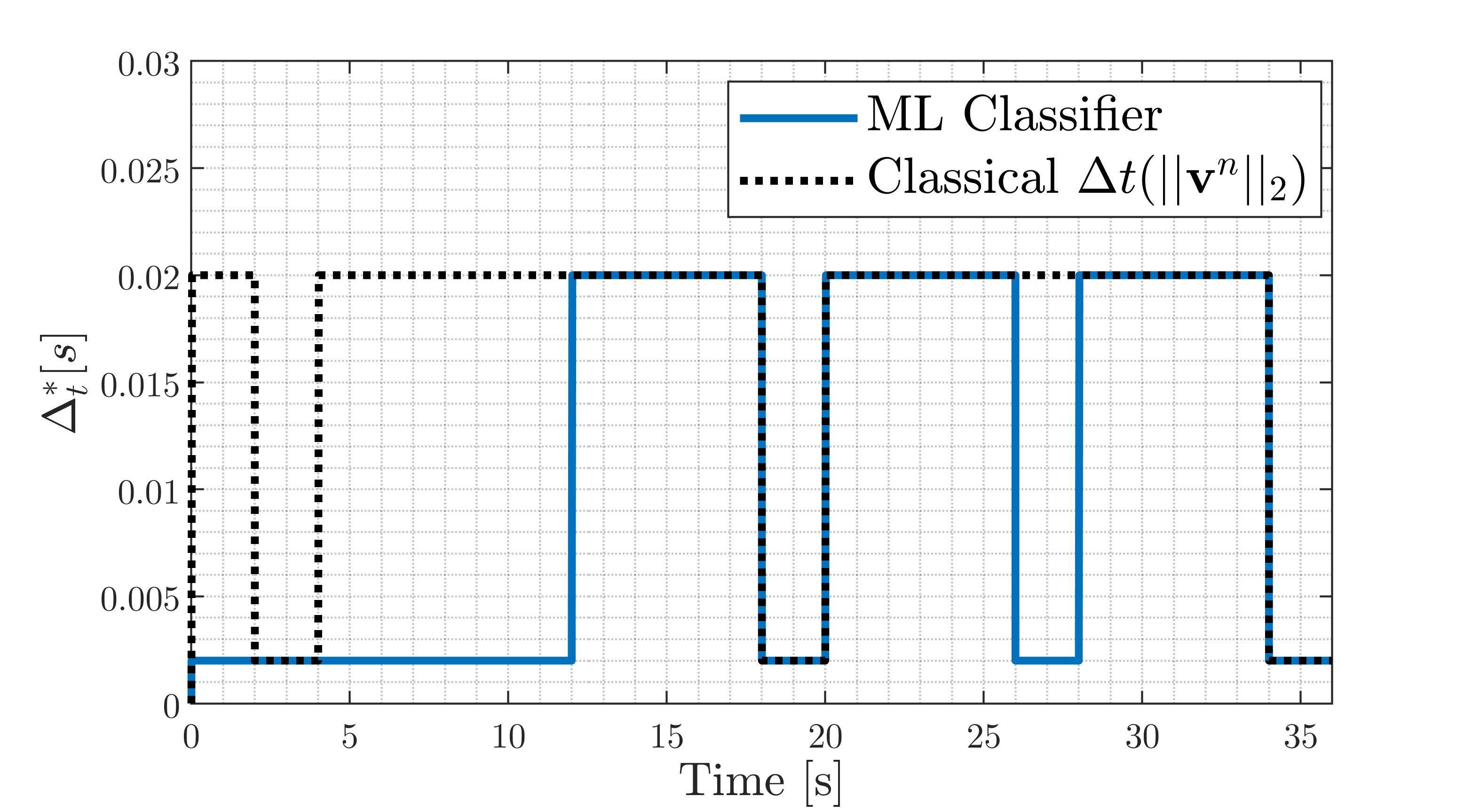}}
\caption{INS/GNSS experiment with sub-optimal step size, $\Delta t^*$, based on the ML classifier and classical $\Delta t\left( {{{\left\| {{{\bf{v}}^n}} \right\|}_2}} \right)$ as a function of time  for a duration of 35 seconds.}
\label{[Fig10}
\end{figure}

\section{CONCLUSIONS}
A proper choice of the step size is important for implementing velocity aided INS. The step size depends on various navigation parameters. In real life, these parameters and their behavior in the dynamic environment are partly known. A novel ML-based scheme to adaptively tune an appropriate step size, together with the es-EKF implementation was proposed. According to this scheme, the designer should set an averaged velocity error bound, where the ML classifier predicts the sub-optimal step size to provide the navigation solution without exceeding this bound, in real-time scenarios. Extensive simulations and a field experiment demonstrated the efficiency of this methodology in commonly used vehicle tracking problems. The proposed scheme minimized the computational load with minimum influence on velocity estimate error. The scheme was validated using two simulations and field experiment. There, the relationship between the velocity RMSE and the IMU step size was presented. In the INS/DVL, INS/GNSS and field experiment, we measured the velocity only and found that we can use lower number of iterations, and by that minimize computational load, with a sufficient velocity RMSE by applying the suggested scheme. In this work, for demonstrative proposes, only two different step size were examined, yet the proposed approach can be easily elaborated to more different sizes. In addition, the goal of this work was focused on velocity aided INS, however the proposed approach can be used with any other aiding sensors and to any other platform.

 \section*{Acknowledgement}
BO was supported by The Maurice Hatter Foundation.

\appendix

\subsection{INS Equations of Motion}
The INS equations of motion include the rate of change of the position, velocity, and the transformation between the navigation and body frame, as shown in Fig.12. \\  The position vector is given by
\begin{equation}
{{\bf{p}}^n} = {\left[ {\begin{array}{*{20}{c}}
\phi &\lambda &h
\end{array}} \right]^T} \in {\mathbb{R}^{3 \times 1}},
\end{equation}
where $\phi$ is the latitude, $\lambda$ is the longitude, and $h$ is the altitude. The velocity vector is Earth referenced and expressed in the North-East-Down (NED) coordinate system:
\begin{equation}
{{\bf{v}}^n} = {\left[ {\begin{array}{*{20}{c}}
{{v_N}}&{{v_E}}&{{v_D}}
\end{array}} \right]^T} \in \mathbb{R}^{3 \times 1},
\end{equation}
where $v_N,v_E,v_D$ denote the velocity vector components in north, east, and down directions, respectively. The rate of change of the position is given by \cite{farrell2008aided}
\begin{equation}
{{{\bf{\dot p}}}^n} = \left[ {\begin{array}{*{20}{c}}
{\dot \phi }\\
{\dot \lambda }\\
{\dot h}
\end{array}} \right] = \left[ {\begin{array}{*{20}{c}}
{\frac{{{v_N}}}{{{R_M} + h}}}\\
{\frac{{{v_E}}}{{\cos \left( \phi  \right)\left( {{R_N} + h} \right)}}}\\
{ - {v_D}}
\end{array}} \right],
\end{equation}
where $R_M$ and $R_N$ are the meridian radius and the normal radius of curvature, respectively.
The rate of change of the velocity vector is given by \cite{farrell2008aided}
\begin{equation}
{{{\bf{\dot v}}}^n} = {\bf{T}}_b^n{{\bf{f}}^b} + {{\bf{g}}^n} - \left( {\left[ {\omega _{en}^n \times } \right] + 2\left[ {\omega _{ie}^n \times } \right]} \right){{\bf{v}}^n},
\end{equation}
where ${\bf{T}}_b^n \in \mathbb{R}^{3 \times 3}$ is the transformation matrix from body frame  to the navigation frame. ${{\bf{f}}^b} \in \mathbb{R}^{3 \times 1}$ is the accelerometers vector expressed in the body frame, ${{\bf{g}}^n} \in \mathbb{R}^{3 \times 1}$ is the gravity vector  expressed in the navigation frame. $\omega _{en}^n$ is the angular velocity vector between the earth centered earth fixed (ECEF) frame and the navigation frame. The angular velocity vector between ECEF and the inertial frame is given by $\omega _{ie}^n$ and the rate of change of the transformation matrix is given by \cite{farrell2008aided}
\begin{equation}
{\bf{\dot T}}_b^n = {\bf{T}}_b^n\left( {\left[ {\omega _{ib}^b \times } \right] - \left[ {\omega _{in}^b \times } \right]} \right),
\end{equation}
where $\omega _{ib}^b = {\left[ {\begin{array}{*{20}{c}}
p&q&r
\end{array}} \right]^T} \in \mathbb{R}^{3 \times 1}$ is the angular velocity vector as obtained by the gyroscope and $\omega _{in}^b$ is the angular velocity vector between the navigation frame and the inertial frame expressed in the body frame. The angular velocity between the navigation frame and the inertial frame expressed in the navigation frame is given by $\omega _{in}^n$. The alignment between body frame and navigation frame can be obtained from ${{\bf{T}}_b^n}$, as follows
\begin{equation}
{\bf \varepsilon}=\left[ {\begin{array}{*{20}{c}}
\varphi \\
\theta \\
\psi 
\end{array}} \right] = \left[ {\begin{array}{*{20}{c}}
{atan2\left( {{\bf{T}}{{_n^b}_{31}},{\bf{T}}{{_n^b}_{32}}} \right)}\\
{arccos\left( {{\bf{T}}{{_n^b}_{33}}} \right)}\\
-{atan2\left( {{\bf{T}}{{_n^b}_{13}},{\bf{T}}{{_n^b}_{23}}} \right)}
\end{array}} \right] \in \mathbb{R}^{3 \times 1},
\end{equation}
where $\varphi$ is the roll angle, $\theta$ is the pitch angle, and $\psi$ is the yaw angle. These three angles are called Euler angles.
The system matrix, $\bf F$, is given by
\begin{equation}
{\bf{F}} = \left[ {\begin{array}{*{20}{c}}
{{{\bf{F}}_{vv}}}&{{{\bf{F}}_{v\varepsilon }}}&{{\bf{T}}_b^n}&{{{\bf{0}}_{3 \times 3}}}\\
{{{\bf{F}}_{\varepsilon v}}}&{{{\bf{F}}_{\varepsilon \varepsilon }}}&{{{\bf{0}}_{3 \times 3}}}&{{\bf{T}}_b^n}\\
{{{\bf{0}}_{3 \times 3}}}&{{{\bf{0}}_{3 \times 3}}}&{{{\bf{0}}_{3 \times 3}}}&{{{\bf{0}}_{3 \times 3}}}\\
{{{\bf{0}}_{3 \times 3}}}&{{{\bf{0}}_{3 \times 3}}}&{{{\bf{0}}_{3 \times 3}}}&{{{\bf{0}}_{3 \times 3}}}
\end{array}} \right]
\end{equation}
where ${{\bf{T}}_b^n}$ is calculated by (18), and ${{\bf{F}}_{ij}} \in {{\mathbb{R}}^{3 \times 3}}$ can be found explicitly in the classical literature (see \cite{farrell2008aided,groves2015principles,brown1992introduction}). 

The dynamic matrix terms ${\bf F}_{ij}$ are provided:
\begin{equation}
{{\bf F}_{\varepsilon v}} = \left[ {\begin{array}{*{20}{c}}
0&{\frac{{ - 1}}{{{R_N} + \hat h}}}&0\\
{\frac{1}{{{R_M} + \hat h}}}&0&0\\
0&{\frac{{\tan \left( {\hat \phi } \right)}}{{{R_N} + \hat h}}}&0
\end{array}} \right]
\end{equation}

\begin{equation}
{{\bf{F}}_{v\varepsilon }} = \left[ {\begin{array}{*{20}{c}}
0&{{{\bf{f}}_D}}&{ - {{\bf{f}}_E}}\\
{ - {{\bf{f}}_D}}&0&{{{\bf{f}}_N}}\\
{{{\bf{f}}_E}}&{ - {{\bf{f}}_N}}&0
\end{array}} \right]
\end{equation}
where matrix terms are the specific forces in navigation frame.

\begin{equation}
{{\bf F}_{\varepsilon \varepsilon }} = \left[ {\begin{array}{*{20}{c}}
0&{{\omega _D}}&{ - {\omega _E}}\\
{ - {\omega _D}}&0&{{\omega _N}}\\
{{\omega _E}}&{ - {\omega _N}}&0
\end{array}} \right]
\end{equation}
where,
\begin{equation}
\left[ {\begin{array}{*{20}{c}}
{{\omega _N}}\\
{{\omega _E}}\\
{{\omega _D}}
\end{array}} \right] = \left[ {\begin{array}{*{20}{c}}
{\left( {\dot {\hat \lambda}  + {\omega _{ie}}} \right)\cos \left( {\hat \phi } \right)}\\
{ - \dot {\hat \phi} }\\
{ - \left( {\dot {\hat \lambda}  + {\omega _{ie}}} \right)\sin \left( {{\hat \phi} } \right)}
\end{array}} \right]
\end{equation}
The ${\bf F}_{vv}$ matrix columns are given as follows
\begin{equation}
\begin{array}{l}
{\bf{F}}_{vv}^{\left( 1 \right)} = \left[ {\begin{array}{*{20}{c}}
{\frac{{{{\hat v}_D}}}{{{R_e}}}}\\
{ - \left( {{\omega _D} - {\omega _{ie}}\sin \left( {\hat \phi } \right)} \right)}\\
{2\frac{{{{\hat v}_N}}}{{{R_e}}}}
\end{array}} \right]\\
{\bf{F}}_{vv}^{\left( 2 \right)} = \left[ {\begin{array}{*{20}{c}}
{2{\omega _D}}\\
{\frac{{{{\hat v}_D}}}{{{R_e}}} + \frac{{{{\hat v}_N}}}{{{R_e}}}\tan \left( \hat \phi  \right)}\\
{ - 2{\omega _N}}
\end{array}} \right]\\
{\bf{F}}_{vv}^{\left( 3 \right)} = \left[ {\begin{array}{*{20}{c}}
{ - \frac{{{{\hat v}_N}}}{{{R_e}}}}\\
{{\omega _N} + {\omega _{ie}}\cos \left( {\hat \phi } \right)}\\
0
\end{array}} \right]
\end{array}
\end{equation}

The shaping matrix is given explicitly by 
\begin{equation}
{\bf{G}} = \left[ {\begin{array}{*{20}{c}}
{{\bf{T}}_b^n}&{{{\bf{0}}_{3 \times 3}}}&{{{\bf{0}}_{3 \times 3}}}&{{{\bf{0}}_{3 \times 3}}}\\
{{{\bf{0}}_{3 \times 3}}}&{{\bf{T}}_b^n}&{{{\bf{0}}_{3 \times 3}}}&{{{\bf{0}}_{3 \times 3}}}\\
{{{\bf{0}}_{3 \times 3}}}&{{{\bf{0}}_{3 \times 3}}}&{{{\bf{I}}_{3 \times 3}}}&{{{\bf{0}}_{3 \times 3}}}\\
{{{\bf{0}}_{3 \times 3}}}&{{{\bf{0}}_{3 \times 3}}}&{{{\bf{0}}_{3 \times 3}}}&{{{\bf{I}}_{3 \times 3}}}
\end{array}} \right]
\end{equation}

\subsection{Evaluation Criterions}
\subsubsection{AuC} 
Binary decision problems are commonly evaluated using the receiver operating characteristic (ROC) curve. One of the large advantages of the ROC is its representation capability of accuracy of the test data. The ROC is a plot of sensitivity vs. specificity. These two parameters are also known as:
\begin{equation}
    TPR = \frac{{TP}}{{TP + FN}}
\end{equation}
and,
\begin{equation}
   FPR = \frac{{FP}}{{FP + TN}}
\end{equation}
where TPR is the true positive rate (sensitivity), FPR is the false positive rate (specificity). $P$ is the amount of positive values, $N$ is the amount of negative values, $TP$ is the number of true positive, $TN$ is the number of true negative, $FP$ is the number of false positive (type one error), and $FN$ is the number of false negative (type two error). The AuC, Area under the Curve, is a measure of the two-dimensional area underneath the entire ROC curve. This measure is scale-invariant and classification threshold invariant. Hence, it is a very useful criterion for classification performance evaluation. 
\subsubsection{Accuracy} 
The second criterion used in order to evaluate the proposed models is the accuracy measure, given by:
\begin{equation}
 ACC{\rm{ }} = {\rm{ }}\frac{{TP{\rm{ }} + {\rm{ }}TN}}{{TP{\rm{ }} + {\rm{ }}TN{\rm{ }} + {\rm{ }}FP{\rm{ }} + {\rm{ }}FN}}
\end{equation}
The accuracy is used as a measure of "how well a binary classification test correctly identifies a condition". It compares estimates of pre and post test probability. This is the ratio of the number of true classified examples over the total number of examples.

\subsection{Bi vs. Multi Classification and Regression Formalization}
The major benefit of defining the problem as a bi-classification predictor, is that we minimize the number of “chattering” between many step size values (might lead to unstable filter). Also, using two values, one big 0.04[s] and the other small 0.002[s], presents clearly the computational effort reduction. A confusion matrix of 10 step size classes are presented here, to demonstrate the lower robustness of considering too much classes for this task. This is part of our initial analysis and is not included in the paper. \\
Also, we formulated this problem as a regression task, where, unfortunately, the obtained trained models result in high RMSE respectively to the step size (linear regression RMSE: 0.025, Tree RMSE: 0.016). 
\begin{figure}[h!]
\centering
{\includegraphics[width=0.45\textwidth]{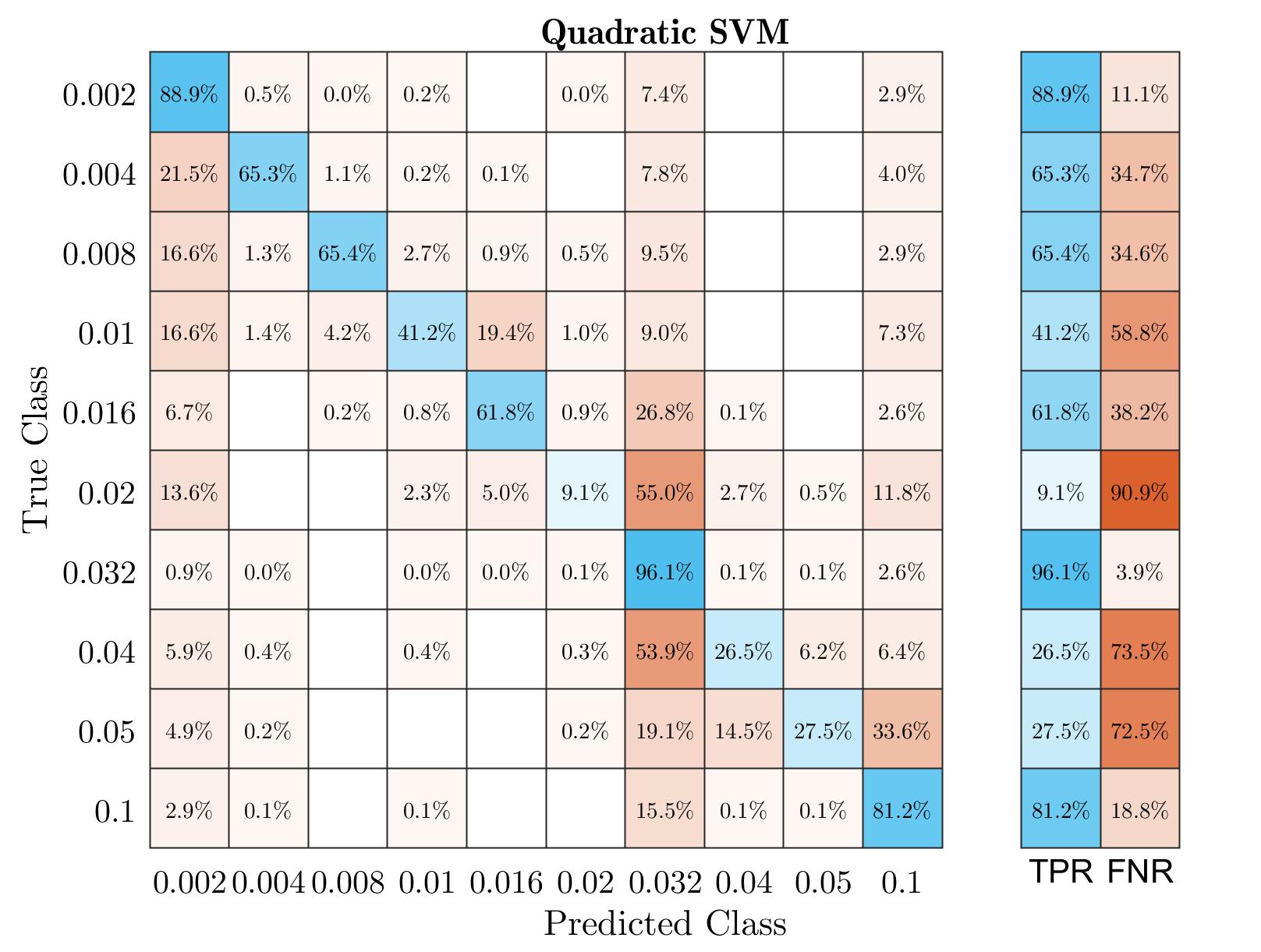}}
\caption{Quadratic SVM classifier for 10 classes of $\Delta t$.}
\label{[Fig10}
\end{figure}

\bibliographystyle{IEEEtran}
\bibliography{IEEEfull}

\begin{IEEEbiography}[{\includegraphics[width=1in,height=1.25in,clip,keepaspectratio]{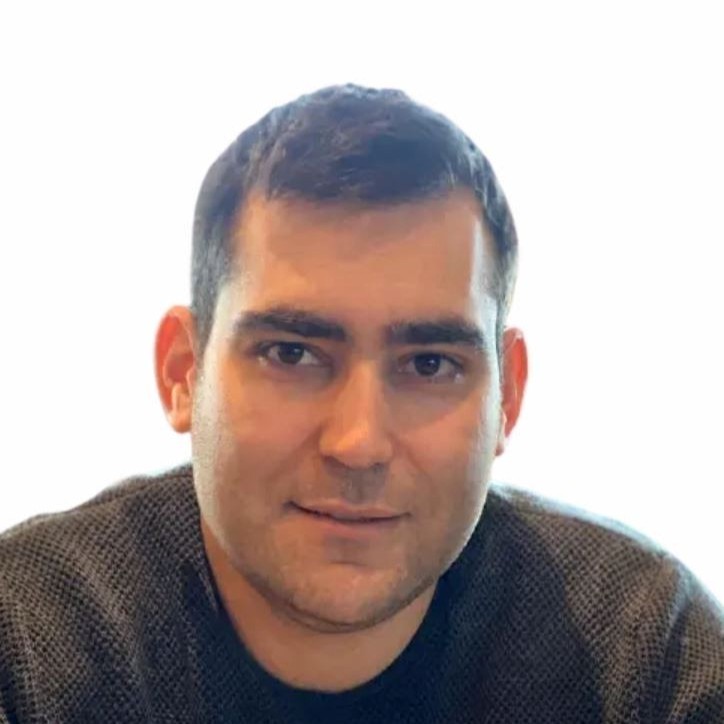}}]{Barak Or} (Member, IEEE) received a B.Sc. degree in aerospace engineering from the Technion–Israel Institute of Technology, Haifa, Israel, a B.A. degree (cum laude) in economics and management, and an M.Sc. degree in aerospace engineering from the Technion–Israel Institute of Technology in 2016 and 2018. He is currently pursuing a Ph.D. degree with the University of Haifa, Haifa. \\
His research interests include navigation, deep learning, sensor fusion, and estimation theory.
\end{IEEEbiography}

\begin{IEEEbiography}[{\includegraphics[width=1in,height=1.25in,clip,keepaspectratio]{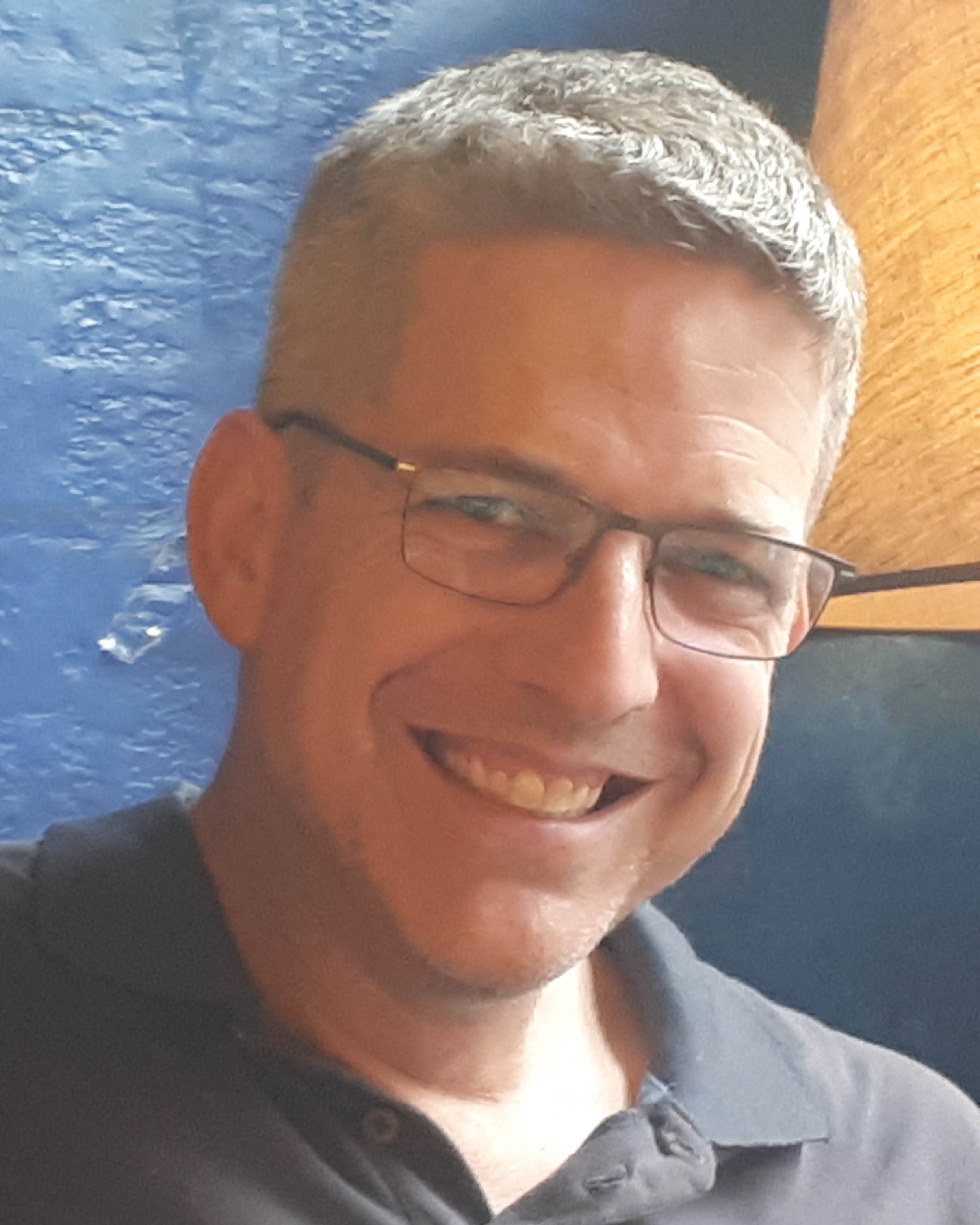}}]{Itzik Klein} (Senior Member IEEE) received B.Sc. and M.Sc. degrees in Aerospace Engineering from the Technion-Israel Institute of Technology, Haifa, Israel in 2004 and 2007, and a Ph.D. degree in Geo-information Engineering from the Technion-Israel Institute of Technology, in 2011. He is currently an Assistant Professor, heading the Autonomous Navigation and Sensor Fusion Lab, at the Hatter Department of Marine Technologies, University of Haifa. His research interests include data driven based navigation, novel inertial navigation architectures, autonomous underwater vehicles, sensor fusion, and estimation theory.
\end{IEEEbiography}

\end{document}